\documentclass[letterpaper,10pt,journal,twoside]{IEEEtran} 
\IEEEoverridecommandlockouts
\let\labelindent\relax  
\usepackage{enumitem}
\setlist[enumerate]{parsep=0pt}
\usepackage{verbatim}
\usepackage[pdftex]{graphicx}
\usepackage{epsfig}
\usepackage{amsmath,amssymb,amsfonts,bm,amscd} 
\usepackage{mathrsfs}
\usepackage{setspace}
\usepackage{fancyhdr}
\usepackage{balance}
\usepackage{psfrag}
\usepackage{cite}
\usepackage{tikz}
\usepackage[utf8]{inputenc}
\usepackage[ruled,vlined,linesnumbered,noresetcount]{algorithm2e}
\usepackage{pgfplots} 
\usepackage{pgfgantt}
\usepackage{pdflscape}
\usepackage[short]{optidef}
\pgfplotsset{compat=newest} 
\pgfplotsset{plot coordinates/math parser=false}
\usepackage[hidelinks]{hyperref}

\usepackage[caption=false,font=footnotesize]{subfig}

\newlength\fwidth

\newcommand{\norm}[1]{\left\Vert#1\right\Vert}

\newcommand{\RR}{{\mathbb R}}

\newcommand{\bhat}[1]{\mathbf{\hat{\text{$#1$}}}}

\newcommand{\btilde}[1]{\mathbf{\tilde{\text{$#1$}}}}

\title{
Trajectory Generation for Multiagent Point-To-Point Transitions via Distributed Model Predictive Control
}

\author{Carlos E. Luis$^{1}$ and Angela P. Schoellig$^{1}$%
	\thanks{Manuscript received: September 10, 2018; Revised December 1, 2018; Accepted December 15, 2018.}
	\thanks{This paper was recommended for publication by Editor Nak Young Chong upon evaluation of the Associate Editor and Reviewers' comments. This work was supported by NSERC research and equipment grants (RTI 2018-00847, CRDPJ 528161-18, CREATE 466088), and the CFI JELF/ORF grant \#33000.} 
	\thanks{$^{1}$Carlos E. Luis and Angela P. Schoellig are with the Dynamic Systems Lab \href{www.dynsyslab.org}{(www.dynsyslab.org)}, Institute for Aerospace Studies, University of Toronto, Canada. E-mails:
		{\tt\footnotesize carlos.luis@robotics.utias.utoronto.ca}, {\tt\footnotesize schoellig@utias.utoronto.ca}}%
	\thanks{Digital Object Identifier (DOI): see top of this page.}
}

\usepackage{xcolor,colortbl}

\newcommand{\todo}[1]{\textcolor{black}{#1}}

\usepackage{bm}

\begin{document}
	
\markboth{IEEE Robotics and Automation Letters. Preprint Version. Accepted December, 2018}
{Luis \MakeLowercase{\textit{et al.}}: Multiagent Trajectory Generation via Distributed Model Predictive Control}

\setlength{\parskip}{0pt}
\maketitle

\begin{abstract}
This paper introduces a novel algorithm for multiagent offline trajectory generation based on distributed model predictive control (DMPC). \todo{Central to the algorithm's scalability and success is the development of an on-demand collision avoidance strategy.} By predicting future states and sharing this information with their neighbours, the agents are able to detect and avoid collisions while moving towards their goals. The proposed algorithm can be implemented in a distributed fashion and reduces the computation time by more than 85\% compared to previous optimization approaches based on sequential convex programming (SCP), while only having a small impact on the optimality of the plans. The approach was validated both through extensive simulations and experimentally with teams of up to 25 quadrotors flying in confined indoor spaces.
\end{abstract}



\setcounter{section}{0}
\section{Introduction}
\label{sec:introduction}
\IEEEPARstart{G}{enerating} collision-free trajectories when dealing with multiagent systems is a safety-critical task. In missions that require cooperation of multiple agents, such as warehouse management~\cite{guizzo2008three}, we often must safely drive agents from their current locations to a set of final positions. Solving this task, known as multiagent point-to-point transition, is therefore an integral part of any robust multiagent system.

There are two main variations of the multiagent point-to-point transition problem: the labelled and the unlabelled agent problem. In the former, each agent has a fixed final position that cannot be exchanged with another agent \cite{schouwenaars2001mixed, augugliaro2012generation}; in the latter, the algorithm is free to assign the goals to the agents, as to ease the complexity of the transition problem \cite{turpin2012decentralized}. This paper focuses on the labelled agent problem. 

A common approach is to formulate this as an optimization problem. One of the first techniques developed relied on Mixed Integer Linear Programming (MILP), modelling collision constraints with binary variables \cite{schouwenaars2001mixed}. This method is computationally expensive and not suited for large groups of agents.

\begin{figure}[t]
	\vspace{1.2ex}
	\centering
	\includegraphics[width=0.45\textwidth]{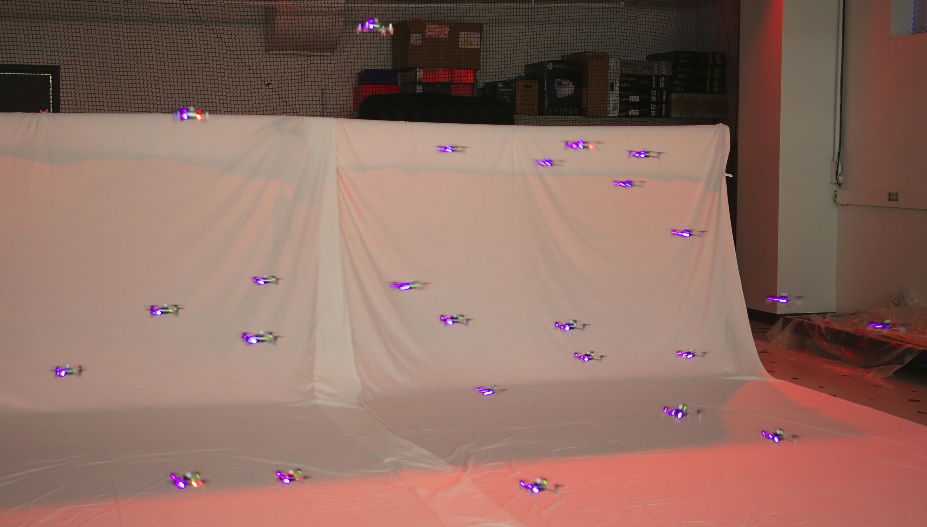}
	\caption{A group of 25 Crazyflie 2.0 quadrotors performing a point-to-point transition using our distributed model predictive control (DMPC) algorithm. A video of the performance is found at
\href{http://tiny.cc/dmpc-swarm}{{\tt http://tiny.cc/dmpc-swarm}}.}
	\label{fig:30_random}
\end{figure}

More recently, Sequential Convex Programming (SCP) \cite{boyd2008sequential} has been used to achieve faster computation compared to MILP. In \cite{augugliaro2012generation}, SCP is used to compute optimal-energy trajectories for quadrotor teams. Although useful for small teams, the algorithm does not scale well with the number of agents. A decoupled version of that algorithm was proposed in \cite{chen2015decoupled,robinson2018efficient}, which provides better scalability at the cost of suboptimal solutions. However, the required decoupling leads to a sequential greedy strategy (i.e., turning agent trajectories previously solved for into obstacles for subsequent agents) with decreased success rate as the number of agents increases.

Discrete approaches divide the space into a grid and use known discrete search algorithms \cite{preiss}, limiting the initial and final locations to be vertices of the underlying grid. Other approaches combine optimization techniques and predefined behaviours to manage collisions in 2D \cite{tang2018complete}. 


Distributed optimization approaches can effectively include pair-wise distance constraints \cite{bhattacharya2011distributed}. Furthermore, the computational effort is distributed among the agents and therefore reduced compared to centralized approaches. Optimal reciprocal collision avoidance (ORCA) leverages velocity obstacles to guarantee collision-free trajectories for holonomic \cite{van2011reciprocal} and non-holonomic \cite{alonso2013optimal} agents. While provably safe, the method may be overly conservative by assuming a constant velocity profile over the time horizon. \todo{Techniques based on potential fields have been used for decentralized collision avoidance \cite{rezaee2014decentralized}, but they are susceptible to deadlocks.}

Distributed model predictive control (DMPC) \cite{camponogara2002distributed} has been used in coordination tasks such as formation control \cite{van2017distributed,sayyaadi2017decentralized}, but not explicitly for point-to-point transitions. Particularly interesting are synchronous implementations of DMPC \cite{dai2017distributed}, where the agents simultaneously update their predictions, reducing runtime by parallel computing.

\todo{Previous DMPC approaches achieved collision avoidance by either (1) using compatibility constraints that limit the position deviation of agents between prediction updates \cite{wang2014synthesis} or (2) imposing separating hyperplane constraints between the agents at every time step of the prediction horizon \cite{van2017distributed}. Both strategies are not well suited for transition tasks: strategy (1) drastically reduces the mobility of agents, especially in cluttered environments, while strategy (2) lacks scalability and is overly conservative, as demonstrated in Sec.~\ref{subsec:hardvssoft}. In contrast, inspired by the incremental inclusion of \textit{all} collision constraints over an \textit{infinite} horizon proposed in \cite{chen2015decoupled}, we introduce \textit{on-demand collision avoidance} in a DMPC framework, where we detect and resolve only the \textit{first} collision in the \textit{finite} prediction horizon, reducing computation time and increasing the success rate for transition tasks. Our method is further enhanced by the use of soft collision constraints, as in \cite{papen2017collision}.}
	



The key contributions of this paper are three-fold: we introduce a novel on-demand collision avoidance strategy for DMPC, present a fast DMPC algorithm for multiagent point-to-point transitions, and provide a thorough empirical analysis of our method via simulations and real quadrotor experiments, as well as comparisons to existing approaches. To the best of our knowledge, our method is the first to be fast enough for midflight trajectory generation with 25 drones (computations are done upon request during flight).

The rest of the paper is organized as follows: Section~II states the problem. Section III  introduces the optimization formulation to solve it. The algorithm is presented in Section IV and demonstrated in simulation (Section V) and experiments with a swarm of quadrotors (Section VI).

\section{Problem Statement}
\label{sec:formulation}
The goal is to generate collision-free trajectories that drive $N$ agents from initial to final locations within a given amount of time, subject to state and actuation constraints. We aim to generate such trajectories offline and execute them with our experimental platform, the Crazyflie 2.0 quadrotor.

\subsection{The Agents}
The agents are modeled as unit masses in $\RR^3$, with double integrator dynamics. This simplified model of a quadrotor with an underlying position controller is used to achieve faster computations. Higher-order models can be accommodated with minimum modifications in what follows.  We use $\textbf{p}_i[k]$, $\textbf{v}_i[k]$, $\textbf{a}_i[k]$ to represent the discretized \todo{$x$, $y$, $z$} position, velocity and accelerations of agent $i$ at time step $k$, where accelerations are the inputs. With a discretization step $h$, the dynamic equations are given by
\begin{align}
\textbf{p}_i[k+1] &= \textbf{p}_i[k] + h\textbf{v}_i[k] + \frac{h^2}{2}\textbf{a}_i[k], \label{eqn:dyn1} \\
\textbf{v}_i[k+1] &= \textbf{v}_i[k] + h\textbf{a}_i[k]. \label{eqn:dyn2}
\end{align}
\subsection{Constraints}
We constrain the motion of the agents to match the physics of the vehicle. First, the agents have limited actuation, which bounds its minimum and maximum acceleration,
\begin{equation}
\label{eqn:acc}
\textbf{a}_{\min} \leq \textbf{a}_i[k] \leq \textbf{a}_{\max}.
\end{equation}

Secondly, the agents must remain within a volume (e.g., an indoor flying arena). We impose:
\begin{equation}
\label{eqn:pos}
\textbf{p}_{\min} \leq \textbf{p}_i[k] \leq \textbf{p}_{\max}.
\end{equation}

\subsection{Collision Avoidance}
The collision avoidance constraint is designed such that the agents safely traverse the environment. In the case of quadrotors, aerodynamic effects from neighbouring agents may lead to crashes. Thus, we model the collision boundary for each agent as an ellipsoid elongated along the vertical axis to capture the downwash effect of the agents' propellers, similar to \cite{preiss}. The collision constraint between agents $i$ and $j$ is defined using a scaling matrix $\bm{\Theta}$,
\begin{equation}
\label{eqn:hard_coll2}
\norm{\bm{\Theta}^{-1}\left(\textbf{p}_i[k]-\textbf{p}_j[k]\right) }_n \geq r_{\min},
\end{equation}
where $n$ is the degree of the ellipsoid ($n=2$ is a usual choice) and $r_{\min}$ is the minimum distance between agents in the xy plane. The scaling matrix $\bm{\Theta}$ is defined as ${\bm{\Theta} = \text{diag}(a,b,c)}$. We choose $a=b=1$ and $c>1$. Thus, the required minimum distance in the vertical axis is $r_{z,\min} = cr_{\min}$. \todo{Note that the constraint in (\ref{eqn:hard_coll2}) checks whether agent $j$ (or $i$), modelled as a 3D point, is inside an ellipsoid centered around agent $i$ (or $j$).}

\section{Distributed Model Predictive Control}
The problem formulated in Sec.~\ref{sec:formulation} can be translated into an optimization problem. In single-agent standard model predictive control (MPC), an optimization problem is solved at each time step, which finds an optimal input sequence over a given prediction horizon based on a model that describes the agent's dynamics. The first input of that sequence is applied to the real system and the resulting state is measured, which is the starting point for the next optimization problem. In an offline planning scenario such as ours, we do not measure the agent's state after applying an input (since there is no physical agent yet), instead we apply the input directly to the model to compute the next step of the generated trajectory. The same procedure is repeated until the whole trajectory is generated. This methodology can be applied in a distributed fashion, where each agent executes the iterative optimization to generate trajectories, but with the possibility of sharing information with neighbouring agents.

\subsection{The Synchronous Algorithm}
Our approach is based on synchronous DMPC, where the agents share their previously predicted state sequence with their neighbours before simultaneously solving the next optimization problems. At every discrete-time index $k_t$, each agent  simultaneously computes a new input sequence over the horizon following these steps:

\begin{enumerate}[wide=0pt] 
	\item Check for future collisions using the latest predicted states of the neighbours, computed at time step $k_t-1$.
	\item Build the optimization problem, including state and actuation constraints, and collision constraints \todo{\textit{only if required}}.
	\item After obtaining the next optimal sequence, the first element is applied to the model and the agents move one step ahead. Future states are predicted over the horizon and shared with the other agents.
\end{enumerate}

\todo{Predicting collisions and including constraints only if needed is the basic idea behind on-demand collision avoidance}. We only include those constraints associated with the first predicted collisions. The process is repeated until all agents reach their desired goals. Below we derive the mathematical setup of the optimization problem.

\subsection{The Agent Prediction Model}
Using the dynamics in (\ref{eqn:dyn1}) and (\ref{eqn:dyn2}), we can develop a linear model to express the agents' states over a horizon of fixed length $K$. First we introduce the notation $\bhat{(\cdot)}[k|k_t]$, which represents the predicted value of $(\cdot)[k_t+k]$ with the information available at $k_t$. In what follows, ${k \in \lbrace 0,\dots,K-1 \rbrace}$ is the discrete-time index of the prediction horizon. The dynamic model of agent $i$ is given by
\begin{equation}
\label{eqn: model}
\begin{bmatrix}
\bhat{\textbf{p}}_i[k+1|k_t]\\
\bhat{\textbf{v}}_i[k+1|k_t]
\end{bmatrix} = \begin{bmatrix}
\textbf{I}_3 & h\textbf{I}_3\\
\bm{0}_3 & \textbf{I}_3
\end{bmatrix} \begin{bmatrix}
\bhat{\textbf{p}}_i[k|k_t] \\
\bhat{\textbf{v}}_i[k|k_t]
\end{bmatrix} + \begin{bmatrix}
(h^2/2)\textbf{I}_3 \\
h \textbf{I}_3
\end{bmatrix}\bhat{\textbf{a}}_i[k|k_t],
\end{equation}
with $\textbf{I}_3$ being a $3\times 3$ identity matrix and $\bm{0}_3$ a $3\times 3$ matrix of zeros. We select the acceleration as the model's input (and variable to optimize). A compact representation is
\begin{equation}
\label{eqn:model_abb}
\bhat{\textbf{x}}_i[k+1|k_t] = \textbf{A}\bhat{\textbf{x}}_i[k|k_t] + \textbf{B}\bhat{\textbf{u}}_i[k|k_t],
\end{equation}
where $\hat{\textbf{x}}_i \in \RR^{6}$, $\textbf{A} \in \RR^{6\times 6} $, $\textbf{B} \in \RR^{6\times 3}$ and $\hat{\textbf{u}}_i \in \RR^{3}$ (model input). Define the initial state at instant $k_t$, $\textbf{X}_{0,i} = \textbf{x}_i[k_t]$. Then we can write the position sequence $\textbf{P}_i \in \RR^{3K}$ as an affine function of the input sequence $\textbf{U}_i \in \RR^{3K}$,
\begin{equation}
\label{eqn:model}
\textbf{P}_i = \textbf{A}_0\textbf{X}_{0,i} + \bm{\Lambda} \textbf{U}_i,
\end{equation}
where $\bm{\Lambda} \in \RR^{3K \times 3K}$ is defined as
\begin{equation}
\bm{\Lambda} = \begin{bmatrix}
\bm{\Psi} \textbf{B} & \bm{0}_3 & \ldots & \bm{0}_3\\
\bm{\Psi} \textbf{AB} & \bm{\Psi} \textbf{B} & \ldots & \bm{0}_3 \\
\vdots & \ddots & \ddots & \vdots \\
\bm{\Psi} \textbf{A}^{K-1}\textbf{B} & \bm{\Psi} \textbf{A}^{K-2}\textbf{B} & \dots & \bm{\Psi} \textbf{B} 
\end{bmatrix},
\end{equation}
with matrix $\bm{\Psi} = \begin{bmatrix}
\textbf{I}_3 & \bm{0}_3
\end{bmatrix}$ selecting the first three rows of the matrix products (those corresponding to the position states). 
Lastly, $\textbf{A}_0 \in \RR^{3K \times 6}$ reflects the propagation of the initial state,
\begin{equation}
\textbf{A}_0 = \begin{bmatrix}
( \bm{\Psi} \textbf{A}) ^\intercal & ( \bm{\Psi} \textbf{A}^2) ^\intercal & \ldots & ( \bm{\Psi} \textbf{A}^K) ^\intercal
\end{bmatrix}^\intercal.
\end{equation}

\subsection{Objective Function}
The objective function that is minimized to compute the optimal input sequence has three main components: trajectory error, control effort and input variation. A similar formulation can be found in \cite{ru2017nonlinear}.

\subsubsection{Trajectory error penalty}
This term drives the agents to their goals. We aim to minimize the sum of errors between the positions at the last $\kappa$ time steps of the horizon and the desired final position $\textbf{p}_{d,i}$. The error term is defined as 
\begin{equation}
e_i = \sum _{k = K - \kappa}^K \norm {\bhat{\textbf{p}}_i[k|k_t]-\textbf{p}_{d,i}}_2.
\end{equation}
This term can be turned into a quadratic cost function in terms of the input sequence using (\ref{eqn:model}),
\begin{equation}
\label{eqn:error}
J_{e,i} = \textbf{U}_i^\intercal(\bm{\Lambda}^{\intercal} \btilde{\textbf{Q}} \bm{\Lambda})\textbf{U}_i - 2(\textbf{P}_{d,i}^\intercal \btilde{\textbf{Q}} \bm{\Lambda} -  \left( \textbf{A}_0\textbf{X}_{0,i}\right) ^\intercal \btilde{\textbf{Q}} \bm{\Lambda})\textbf{U}_i,
\end{equation}
where $\btilde{\textbf{Q}} \in \RR^{3K \times 3K} $ is a positive definite and block-diagonal matrix that weights the error at each time step. A value of $\kappa = 1$ leads to $\btilde{\textbf{Q}} = \text{diag}(\bm{0}_3, \dots, \textbf{Q})$ with matrix $\textbf{Q} \in \RR^{3 \times 3}$ chosen as a diagonal positive definite matrix. Higher values of $\kappa$ lead to more aggressive agent behaviour with agents moving faster towards their goals, but may also lead to overshooting at the target location.  

\subsubsection{Control effort penalty}
We also aim to minimize the control effort using the quadratic cost function
\begin{equation}
\label{eqn:input}
J_{u,i} = \textbf{U}_i^\intercal \btilde{\textbf{R}} \textbf{U}_i.
\end{equation}
Similarly, $\btilde{\textbf{R}} \in \RR^{3K \times 3K} $ is positive definite and block-diagonal, $\btilde{\textbf{R}} = \text{diag}(\textbf{R}, \dots, \textbf{R})$, where $\textbf{R} \in \RR^{3 \times 3}$ weights the penalty on the control effort.

\subsubsection{Input variation penalty}
This term is used to minimize variations of the acceleration, leading to smooth input trajectories. We define the quadratic cost
\begin{equation}
\label{eqn:delta}
\delta_i = \sum_{k=0}^{K-1}\norm{\bhat{\textbf{u}}_i[k|k_t]-\bhat{\textbf{u}}_i[k-1|k_t]}_2.
\end{equation}
To transform (\ref{eqn:delta}) into a quadratic form, first we define a matrix $\bm{\Delta} \in \RR^{3K \times 3K}$,
\begin{equation}
\bm{\Delta} = \begin{bmatrix}
\textbf{I}_3 & \bm{0}_3 & \bm{0}_3 & \ldots & \bm{0}_3 & \bm{0}_3 \\
-\textbf{I}_3 & \textbf{I}_3 & \bm{0}_3 & \dots & \bm{0}_3 & \bm{0}_3 \\
\bm{0}_3 & -\textbf{I}_3 & \textbf{I}_3 & \ldots & \bm{0}_3 & \bm{0}_3 \\
\vdots & \ddots & \ddots & \ddots & \vdots & \vdots\\
\bm{0}_3& \bm{0}_3 & \bm{0}_3 & \ldots & -\textbf{I}_3 & \textbf{I}_3
\end{bmatrix},
\end{equation}
and introduce the vector $\textbf{U}_{i\ast} \in \RR^{3K}$ to include the term ${\textbf{u}_i[k_t-1]}$ (previously applied input),
\begin{equation}
\textbf{U}_{i\ast} = \begin{bmatrix}
\textbf{u}_i[k_t-1]^\intercal & \bm{0}_{3\times 1}^\intercal & \ldots & \bm{0}_{3\times 1}^\intercal
\end{bmatrix}^\intercal.
\end{equation}
Finally, we write (\ref{eqn:delta}) in quadratic form as
\begin{equation}
\label{eqn:var}
J_{\delta,i} = \textbf{U}_i^\intercal (\bm{\Delta}^\intercal \btilde{\textbf{S}} \bm{\Delta}) \textbf{U}_i - 2(\textbf{U}_{i\ast}^\intercal \btilde{\textbf{S}} \bm{\Delta} )\textbf{U}_i,
\end{equation}	
where $\btilde{\textbf{S}} \in \RR^{3K \times 3K}$ is positive definite and block-diagonal, defined as $\btilde{\textbf{S}} = \text{diag}(\textbf{S}, \dots, \textbf{S})$,  where $\textbf{S} \in \RR^{3 \times 3}$ weights the penalty on control variation. The cost function $\mathcal{J}_i$ is obtained by adding together (\ref{eqn:error}), (\ref{eqn:input}) and (\ref{eqn:var}),
\begin{equation}
\mathcal{J}_i(\textbf{U}_i) = J_{e,i} + J_{u,i} + J_{\delta,i}
\end{equation}	
\subsection{Physical Limits}
When computing the input sequence over the horizon, the agents must satisfy constraints (\ref{eqn:acc}) and (\ref{eqn:pos}). Define $\textbf{P}_{\min}, \textbf{P}_{\max}, \textbf{U}_{\min}, \textbf{U}_{\max} \in \RR^{3K}$ to be
\begin{equation}
\begin{aligned}
\textbf{P}_{\min} = [\textbf{p}_{\min}^\intercal \ldots \textbf{p}_{\min}^\intercal]^\intercal; \quad  \textbf{P}_{\max} = [\textbf{p}_{\max}^\intercal \ldots \textbf{p}_{\max}^\intercal]^\intercal \\
\textbf{U}_{\min} = [\textbf{a}_{\min}^\intercal \ldots \textbf{a}_{\min}^\intercal]^\intercal; \quad  \textbf{U}_{\max} = [\textbf{a}_{\max}^\intercal \ldots \textbf{a}_{\max}^\intercal]^\intercal.
\end{aligned}
\end{equation}
The physical limits are formulated as
\begin{equation}
\begin{aligned}
\label{eqn:inequality}
\textbf{P}_{\min} - \textbf{A}_0\textbf{X}_{0,i} &\leq \bm{\Lambda} \textbf{U}_i \leq \textbf{P}_{\max} - \textbf{A}_0\textbf{X}_{0,i} \\
\textbf{U}_{\min} &\leq  \textbf{U}_i \leq \textbf{U}_{\max}.
\end{aligned}
\end{equation}
\todo{Lastly, we can vertically stack both inequality constraints in (\ref{eqn:inequality}) to obtain a single expression: $\textbf{A}_{\text{in}}\textbf{U}_i \leq \textbf{b}_{\text{in}}$.}

\subsection{Convex Optimization Problem, No Collision Case}
If agent $i$ does not detect any future collisions, then it updates its input sequence by solving:
\begin{mini}|l|
	{\textbf{U}_i}{\mathcal{J}_i(\textbf{U}_i)}{}{}
	{\label{eqn:convex}}{}
	\addConstraint{\textbf{A}_{\text{in}}\textbf{U}_i}{\leq \textbf{b}_{\text{in}}}.
\end{mini}

The formulation in (\ref{eqn:convex}) results in a quadratic programming problem with $3K$ decision variables and $12K$ inequality constraints, which scales independently of $N$.

\begin{figure*}
	\centering
	\subfloat[$t = 0 \text{s}$]{\label{fig:trans_a}\includegraphics[width=0.23\textwidth]{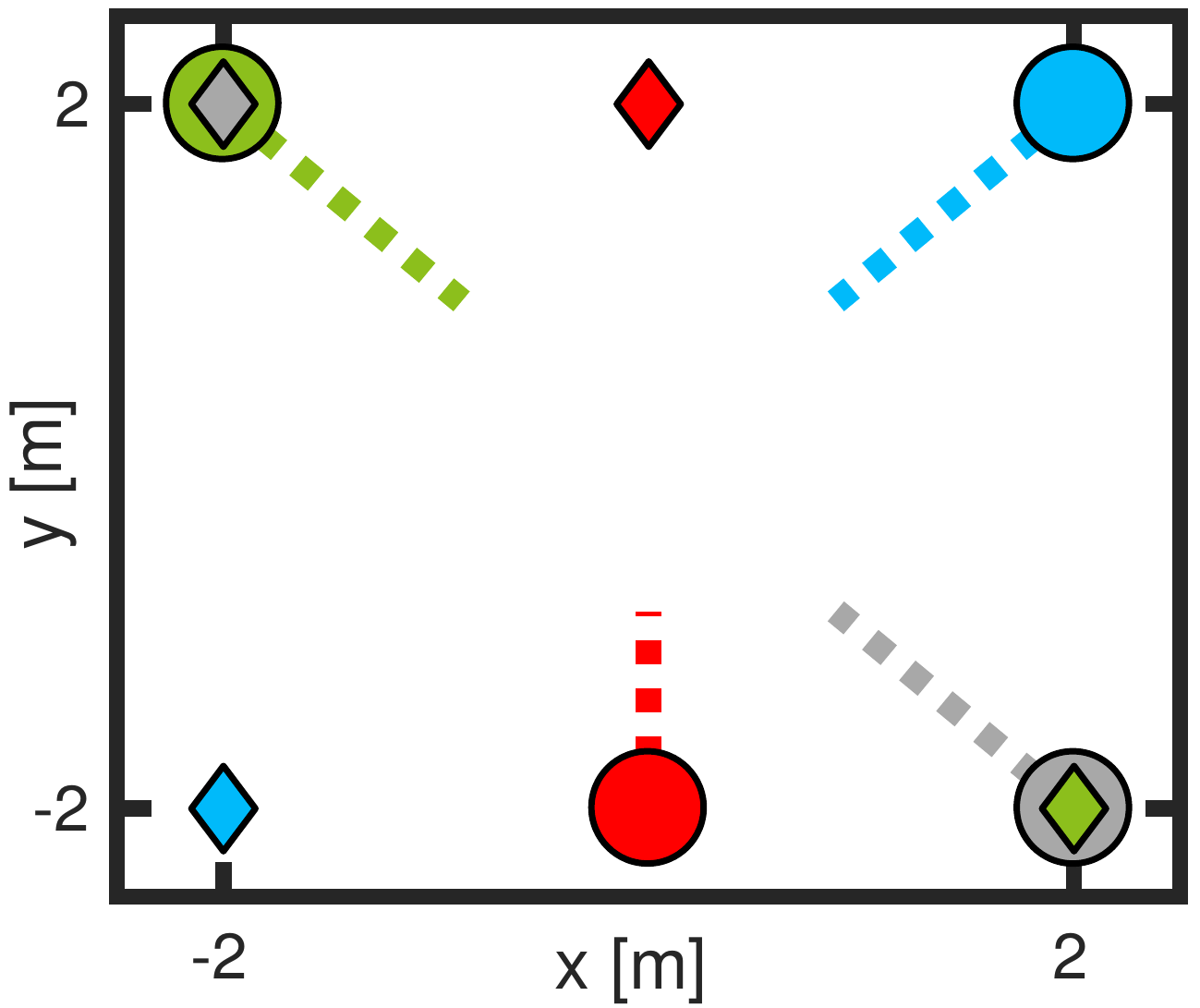}}
	\subfloat[$t = 1 \text{s}$]{\label{fig:trans_b}\includegraphics[width=0.23\textwidth]{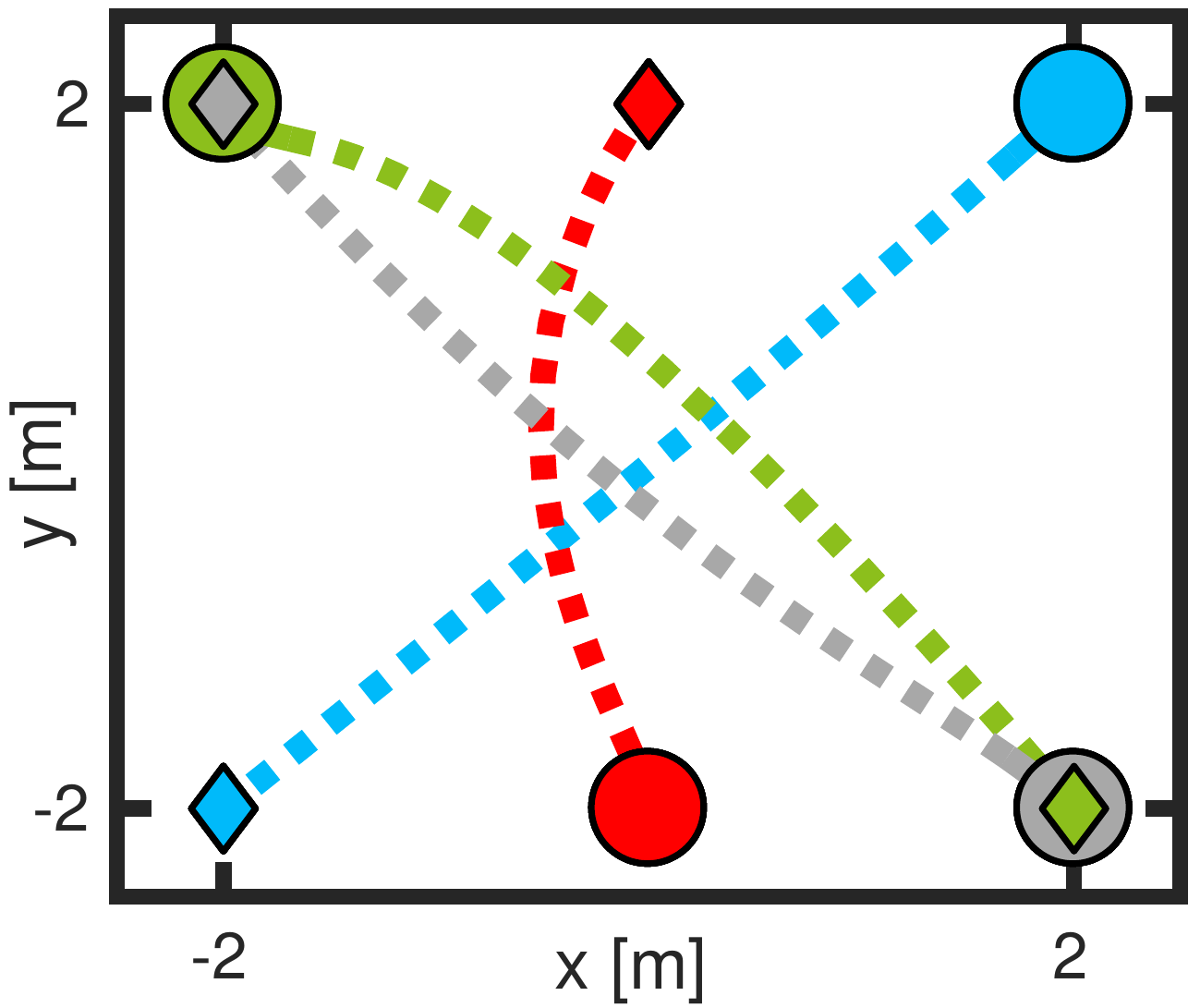}}
	\subfloat[$t = 2 \text{s}$]{\label{fig:trans_c}\includegraphics[width=0.23\textwidth]{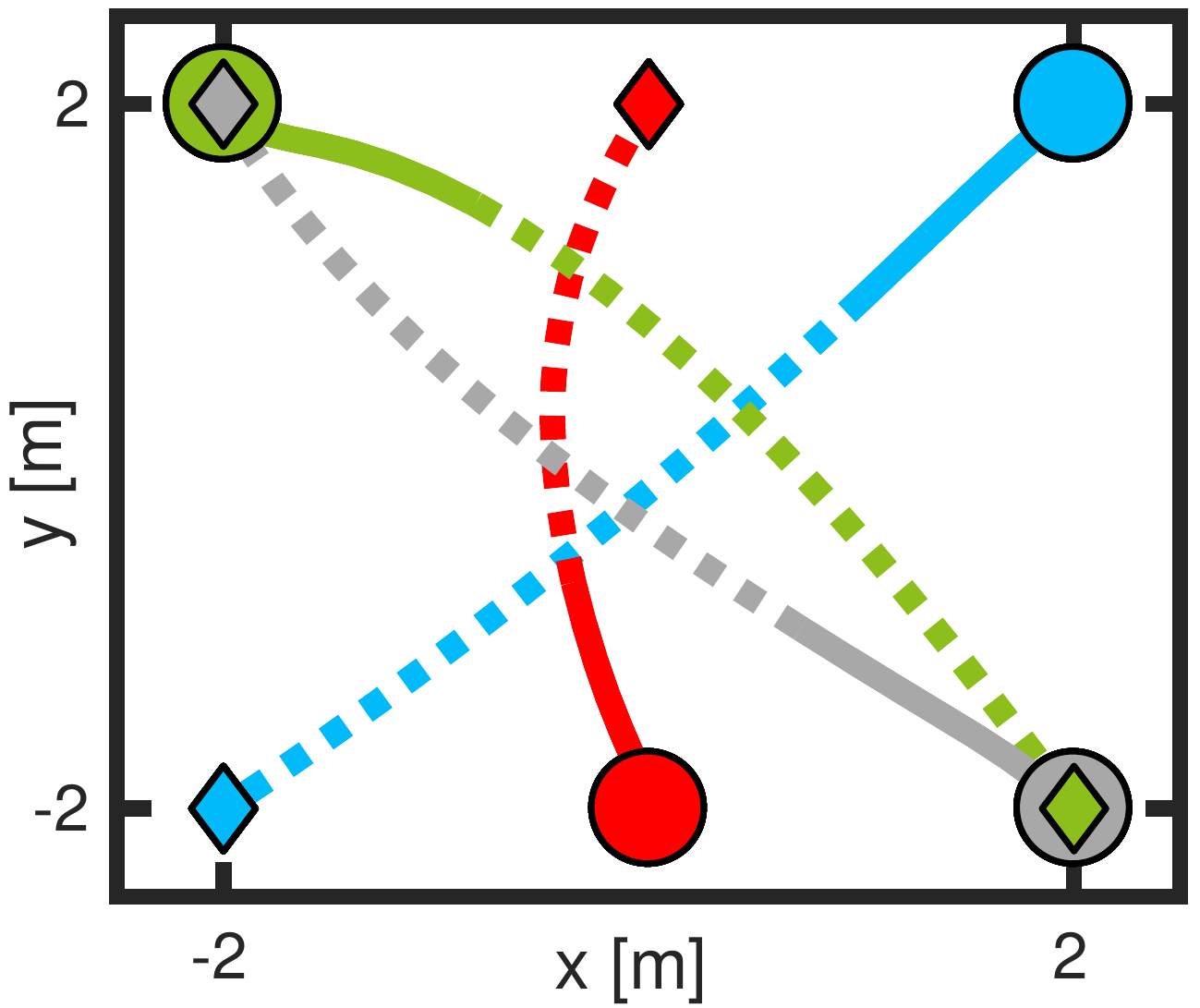}}
	\subfloat[$t = 8.2 \text{s}$]{\label{fig:trans_d}\includegraphics[width=0.23\textwidth]{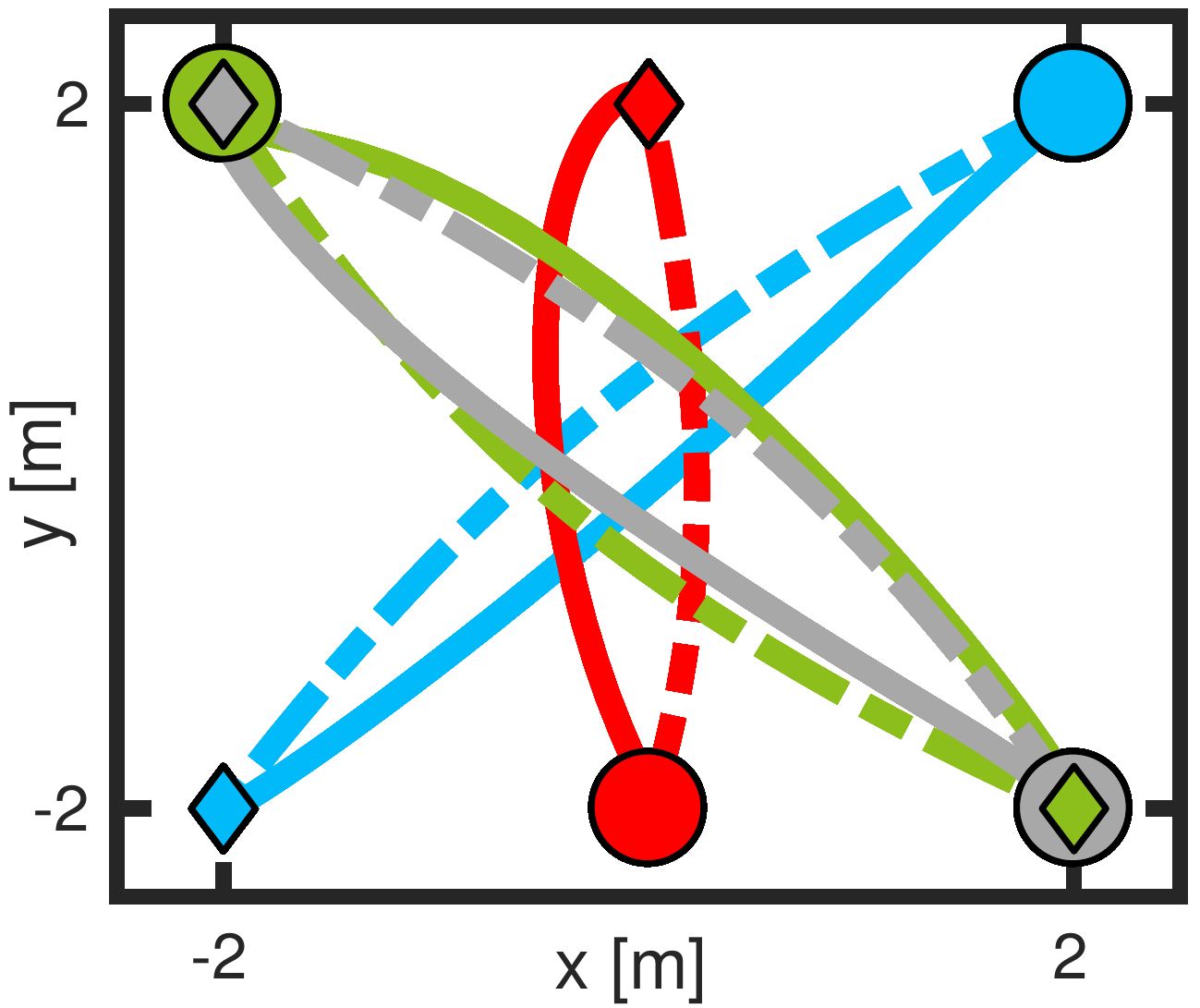}}
	\caption{Four-agent position exchange scenario in 2D solved using Algorithm 1. Circles and diamonds represent initial and final locations, respectively. Dotted lines in \protect\subref{fig:trans_a} - \protect\subref{fig:trans_c} represent the predicted positions over a 3-second horizon, solid lines are the generated trajectories and dashed lines in \protect\subref{fig:trans_d} are the trajectories generated by the centralized approach in  \cite{augugliaro2012generation}. \todo{Using the optimality criteria of the sum of travelled distances by all agents, the distributed plan is only slightly suboptimal when compared to the centralized approach.}}
	\label{fig:four}
\end{figure*}

\subsection{\todo{On-demand Collision Avoidance with Soft Constraints}}
\label{subsec:soft}
The previous formulation is useful for scenarios where the agents can follow straight lines to their goals without colliding. In a more general setting, agents must avoid each other constantly to reach their goals. \todo{To implement on-demand collision avoidance, we leverage the predictive nature of DMPC to detect colliding trajectories and impose constraints to avoid the \textit{first} predicted collision. This strategy differs from \cite{chen2015decoupled} since we do not attempt to incrementally resolve \textit{all} predicted collisions, only the most relevant one}. 

Agent $i$ detects a collision at time step $k_{c,i}$ of the previously considered horizon whenever the inequality
\begin{equation}
\label{eqn:hard_coll}
\xi_{ij} = \norm{\bm{\Theta}^{-1}\left( \bhat{\textbf{p}}_i[k_{c,i}|k_t-1]-\bhat{\textbf{p}}_j[k_{c,i}|k_t-1]\right) }_n \geq r_{\min}
\end{equation}
does not hold with a neighbour $j$. Note that at solving time $k_t$, the agents only have information of the other agents computed at $k_t-1$, meaning that the collision is predicted to happen at time $k_{c,i}+k_t-1$. In what follows, $k_{c,i}$ represents the \textit{first} time step of the horizon where agent $i$ predicts a collision with any neighbour. We include collision constraints with the subset of agents $\Omega_i$ defined as
\begin{equation}
\Omega_i = \lbrace j \in \lbrace 1,\dots,N \rbrace \mid \xi_{ij} < f(r_{\min}), i \neq j \rbrace \nonumber,
\end{equation}
where $f(r_{\min})$ models the radius around the agent, which defines the neighbours to be considered as obstacles when solving the problem. For example, we may include all agents within a radius 3 times bigger than the collision boundary, then $f(r_{\min}) = 3r_{\min}$. \todo{Limiting $\Omega_i$ to be the subset of neighbours within a radius of agent $i$ intends to safely reduce the amount of collision constraints in the optimization.}


If the agent detects collisions, it must include collision constraints to compute the new input sequence. To account for infeasibility issues while solving the optimization problem, we formulate the following relaxed collision constraint:
\begin{equation}
\label{eqn:coll_ellip}
\norm{\bm{\Theta}^{-1}\left( \bhat{\textbf{p}}_i[k_{c,i}-1|k_t]-\bhat{\textbf{p}}_j[k_{c,i}|k_t-1]\right) }_n \geq r_{\min} + \varepsilon_{ij},
\end{equation}
where $\varepsilon_{ij} < 0$ is a new decision variable that relaxes the constraint. Note that at $k_t$, we aim to optimize the value of $\bhat{\textbf{p}}_i[k_{c,i}-1|k_t]$ to satisfy (\ref{eqn:coll_ellip}). The constraint is linearized using a Taylor series expansion about the previous predicted position of agent $i$ at time $k_{c,i} + k_t-1$, namely $\bhat{\textbf{p}}_i[k_{c,i}|k_t-1]$,
\begin{equation}
\label{eqn: bla}
\begin{aligned}
\bm{\nu}_{ij}^\intercal \bhat{\textbf{p}}_i[k_{c,i} | k_t]- \varepsilon_{ij}\xi_{ij} \geq \rho_{ij}
\end{aligned}
\end{equation}
with 
$\bm{\nu}_{ij} = \bm{\Theta}^{-n}( \bhat{\textbf{p}}_i[k_{c,i}|k_t-1]-\bhat{\textbf{p}}_j[k_{c,i}|k_t-1]) ^{n-1} $ and $\rho_{ij} = r_{\min}\xi_{ij} + \xi_{ij}^{n} + \bm{\nu}_{ij}^\intercal \bhat{\textbf{p}}_i[k_{c,i}|k_t-1]$. On the left-hand side of (\ref{eqn: bla}), we note that the constraint is imposed on the position at time $k_t+k_{c,i}$ ($\bhat{\textbf{p}}_i[k_{c,i} | k_t]$), which is one time step after the predicted collision. This choice was made based on an empirical assessment of the algorithm's performance on a wide range of transition scenarios. It was found that by imposing the constraint one time step after the predicted collision, the agents exhibited more preemptive collision avoidance capabilities and were able to complete the transitions faster on average. 

To turn the collision constraint into an affine function of the decision variables, first we augment the previous formulation to include the relaxation variables. Consider $\textbf{E}_i \in \RR^{n_{c,i}}$, with $n_{c,i} = \dim(\Omega_i)$, defined as the stacked vector of all $\varepsilon_{ij}$. We now introduce the augmented decision vector ${\bm{\mathcal{U}}_i \in \RR^{3K+n_{c,i}}}$, obtained by concatenating vectors $\textbf{U}_i$ and $\textbf{E}_i$. The matrices derived above can be easily augmented to account for the augmented decision vector, by completing them with zeros where multiplied with the vector $\textbf{E}_i$. We turn (\ref{eqn: bla}) into an affine function of the decision variables,
\begin{equation}
\label{eqn:ineq}
\bm{\mu}_{ij}^\intercal \bm{\Lambda} \textbf{U}_i - \varepsilon_{ij}\xi_{ij} \geq \rho_{ij} - \bm{\mu}_{ij}^\intercal \textbf{A}_0 \textbf{X}_{0,i},
\end{equation}
where $\bm{\mu}_{ij} \in \RR^{3K}$ is defined as
\begin{equation}
\bm{\mu}_{ij} = \begin{bmatrix}
\bm{0}^\intercal_{3(k_{c,i}-1)\times 1} & \bm{\nu}^\intercal_{ij} & \bm{0}^\intercal_{3(K-k_{c,i})\times 1}
\end{bmatrix}^\intercal.
\end{equation}

By stacking the inequalities in (\ref{eqn:ineq}) for the $n_{c,i}$ colliding neighbours, we obtain the complete collision constraint,
\begin{equation}
\label{eqn:coll_compact}
	\textbf{A}_{\text{coll}} \bm{\mathcal{U}}_i \leq \textbf{b}_{\text{coll}}.
\end{equation}
Additionally, we impose $-\varepsilon_{\max} \leq \varepsilon_{ij} \leq 0$ in order to bound the amount of relaxation allowed. We also consider the following linear and quadratic cost terms to penalize the relaxation on the collision constraint:
\begin{equation}
f_{\varepsilon,i} = \varrho \begin{bmatrix}
\bm{0}^\intercal_{3K\times 1} & \bm{1}^\intercal_{n_{c,i}\times 1}
\end{bmatrix}^\intercal, \textbf{H}_{\varepsilon,i} = \zeta \begin{bmatrix}
\bm{0}_{3K\times 3K} & \bm{0}_{3K \times n_{c,i}} \\
\bm{0}_{n_{c,i} \times 3K} & \textbf{I}_{n_{c,i}}
\end{bmatrix} \nonumber
\end{equation}
where $\varrho$, $\zeta > 0$ are scalar tuning parameters, measuring how much the relaxation is penalized. The augmented cost function in the collision avoidance case is defined as

\begin{equation}
	\mathcal{J}_{\text{aug},i}(\bm{\mathcal{U}}_i) = \mathcal{J}(\textbf{U}_i)  + \bm{\mathcal{U}}_i^\intercal \textbf{H}_{\varepsilon,i} \bm{\mathcal{U}}_i - f_{\varepsilon,i}^\intercal \bm{\mathcal{U}}_i.
\end{equation}

Finally, the convex optimization problem with collision avoidance for agent $i$ is formulated as
\begin{mini}|l|
	{\bm{\mathcal{U}}_i}{\mathcal{J}_{\text{aug},i}(\bm{\mathcal{U}}_i)}{}{}
	{\label{eqn:convex_coll}}{}
	\addConstraint{\textbf{A}_{\text{in,aug}}\bm{\mathcal{U}}_i}{\leq \textbf{b}_{\text{in,aug}}}.
\end{mini}

The subscript `aug' indicates the use of augmented state matrices, as outlined before. \todo{The inequality tuple $(\textbf{A}_{\text{in,aug}}, \textbf{b}_{\text{in,aug}})$ is obtained by vertically stacking the physical limits, the collision constraint and the relaxation variable bounds}. The augmented problem has $3K + n_{c,i}$ decision variables and $12K + 3n_{c,i}$ inequality constraints.
\section{The Algorithm}
\label{sec:problemStatement}
The proposed DMPC algorithm for point-to-point transitions is outlined in Algorithm 1.  It requires as input the initial and desired final locations for $N$ agents ($\textbf{p}_0,\textbf{p}_f$), and outputs the trajectories that complete the transition. Variables $\textbf{p},\textbf{v}$ and $\textbf{a}$ are defined as the concatenation of the transition trajectories for every agent, while $\bm{\Pi}$ is the concatenation of the latest predicted positions for all agents. 

In line 1, every $\bm{\Pi}_i$ is initialized as a line from initial to final location with a constant velocity profile. Each agent's states are initialized to be at the corresponding initial position with zero velocity. The main loop (lines 3-11) repeatedly solves optimization problems for the $N$ agents, building the transition trajectory until they arrive at their goals or a maximum number of time steps is exceeded. \todo{Convergence of the transition (line 10) is declared once \textit{all} the agents are within a small radius of their goals}. Note that for $k_t = 0$, we consider $\textbf{a}_i[-1] = \bm{0}_{3\times1}$. The inner loop (lines 4-9) can be solved either sequentially or in parallel, since there is no data dependency between the problems.

\begin{algorithm}[t]
	\SetKwData{Left}{left}\SetKwData{This}{this}\SetKwData{Up}{up}
	\SetKwFunction{Union}{Union}\SetKwFunction{FindCompress}{FindCompress}
	\SetKwInOut{Input}{Input}\SetKwInOut{Output}{Output}
	\Input {Initial and final positions}
	\Output {Position, velocity and acceleration trajectories}
	
	$[\bm{\Pi},\textbf{x}[0]] \leftarrow$ InitAllPredictions($\textbf{p}_0,\textbf{p}_f$)
	
	$k_t \leftarrow 0 $, AtGoal $\leftarrow$ false
	
	\While{not AtGoal and $k_t < K_{\max}$}{
		\ForEach{agent $i = {1,...,N}$}{
			$\bhat{\textbf{a}}_i[k|k_t]\leftarrow$\todo{Build\&SolveQP}$(\textbf{x}_i[k_t],\textbf{a}_i[k_t-1],\bm{\Pi})$
			
			\If{QP feasible}{
				
				$\bhat{\textbf{x}}_i[k+1|k_t] \leftarrow$ GetStates($\textbf{x}_i[k_t],\bhat{\textbf{a}}_i[k|k_t]$)
				
				$\bm{\Pi}_i \leftarrow \bhat{\textbf{p}}_i[k+1|k_t]$ 
				
				$\textbf{x}_i[k_t+1],\textbf{a}_i[k_t] \leftarrow \bhat{\textbf{x}}_i[1|k_t],\bhat{\textbf{a}}_i[0|k_t]$
			}	
		}
		AtGoal $\leftarrow$ CheckGoal($\textbf{p}[k_t]$,$\textbf{p}_f$)
		
		$k_t \leftarrow k_t  +1$
		
	}
	
	\If{AtGoal}{
		$[\textbf{p},\textbf{v},\textbf{a}] \leftarrow$ ScaleTrajectory($\textbf{p},\textbf{v},\textbf{a},\norm{\textbf{a}_{\max}}$)
		
		$[\textbf{p},\textbf{v},\textbf{a}] \leftarrow$ Interpolate($\textbf{p},\textbf{v},\textbf{a},T_s$)

		\todo{CheckCollisions($\textbf{p}$, $r_{\min} - \varepsilon_{\text{check}}$)}
	}
	\KwRet {\normalfont$[\textbf{p},\textbf{v},\textbf{a}]$}
	\caption{DMPC for Point-to-Point Transitions}
\end{algorithm}

To build and solve the corresponding QP (line 5), first we check for predicted collisions over the horizon, as described in Sec.~\ref{subsec:soft}. If no collisions are detected, we solve the reduced problem in (\ref{eqn:convex}), otherwise we solve the collision avoidance problem in (\ref{eqn:convex_coll}). If the optimizer finds a solution to the QP, then we can propagate the states using (\ref{eqn:model_abb}) and obtain the predicted position and velocity over the horizon (lines 6-9). Lastly, if a solution for the transition was found, we interpolate the solution with time step $T_s$ to obtain a higher resolution trajectory. An optional step is to scale the solution, as suggested in \cite{chen2015decoupled}, to push the accelerations to the maximum allowed. \todo{Finally, in line 15 we perform a collision check by verifying that  ${\norm{\bm{\Theta}^{-1}\left(\textbf{p}_i[k_t]-\textbf{p}_j[k_t]\right) }_n \geq r_{\min} - \varepsilon_{\text{check}}}$ holds for every $i$, $j$ and $k_t$ of the \textit{interpolated} solution. The value of $\varepsilon_{\text{check}} \geq \varepsilon_{\max}$ is user-defined and must reflect the safety limit of the physical agents, such that the algorithm can decide whether the solution is safe to execute or not.  If the solution passes all sanity checks, then the algorithm is deemed successful, otherwise an empty solution is returned.}

\subsection{Example Scenario}
To illustrate how DMPC manages colliding trajectories, Fig.~\ref{fig:four} shows a transition problem for four agents in the plane. Initially, as shown in Fig.~\ref{fig:trans_a}, the agents follow a direct path towards their desired final locations. In Fig.~\ref{fig:trans_b}, collisions are detected and considered in the optimization problem. After a few time steps, the agents obtain the non-colliding plan seen in Fig.~\ref{fig:trans_c}. The trajectories generated with a centralized approach are quite different than the DMPC trajectories, as shown in Fig.~\ref{fig:trans_d}. However, the sum of travelled distance of all agents is fairly similar in both cases, with only a 1.7\% increase for the distributed approach.

\subsection{\todo{Limitations and Associated Mitigation Strategies}}
\label{subsec:repair}
\todo{
We now discuss the limitations of the proposed algorithm, along with associated mitigation strategies to overcome them.
\begin{enumerate}[wide=0pt] 
	\item \textbf{Infeasibility}: the optimization problem becomes infeasible when the constraint (\ref{eqn:coll_compact}) cannot be satisfied given the acceleration and relaxation limits. Feasibility of the problem can be guaranteed, however,  by locally increasing the relaxation bound $\varepsilon_{\max}$ until the constraint is satisfied. In line 5 of Algorithm 1 we apply this technique to ensure recursive feasibility of the problem. The variable $\varepsilon_{\max}$ is reset to its original value once a solution is found. 
	\item \textbf{Collisions}: the use of on-demand collision avoidance with soft constraints does not guarantee collision-free trajectories. The use of soft constraints may lead to partial violations of the collision constraints along the trajectory. Moreover, since the trajectory is specified in discrete-time, there may be collisions occurring between time steps \cite{augugliaro2012generation}. Higher values of $\varrho$ and $\zeta$ penalize the violation of the collision constraint more, rendering the agents more wary of avoiding collisions.
	\item \textbf{Oscillations and deadlocks}: oscillations occur due to a lack of central coordination, where agents oscillate between possible trajectories to avoid a collision. An agent may get trapped in a local minima where it oscillates indefinitely and never reaches its goal (deadlock). Higher values of $\kappa$ and $\textbf{Q}$ encourage aggressiveness towards reaching the goal.
\end{enumerate}
}
\todo{We observed that oscillations are often present in the predictions of agents, but vanish after a few MPC cycles and do not appear in the generated trajectories. Failure to avoid collisions can be minimized by tuning the cost function appropriately, achieved by a good compromise between aggressiveness towards the goal and penalization of the constraint relaxation.}
%

\section{Simulations}
\label{sec:results}
This section provides a simulation analysis of the DMPC algorithm. Implementation was done in MATLAB 2017a (using a sequential implementation of Algorithm 1) and executed on a PC with an Intel Xeon CPU with 8 cores and 16GB of RAM, running at 3GHz. The agents were modelled based on the Crazyflie 2.0 platform, using $r_{\min} = 0.35\,\text{m}$, ${a_{\max} = -a_{\min} = 1\,\text{m/s}^2}$, and $c = 2$ (to avoid downwash).

\begin{figure}[t]
	\centering
	\subfloat[]{\label{fig:prob_hardsoft}\includegraphics[width=0.25\textwidth]{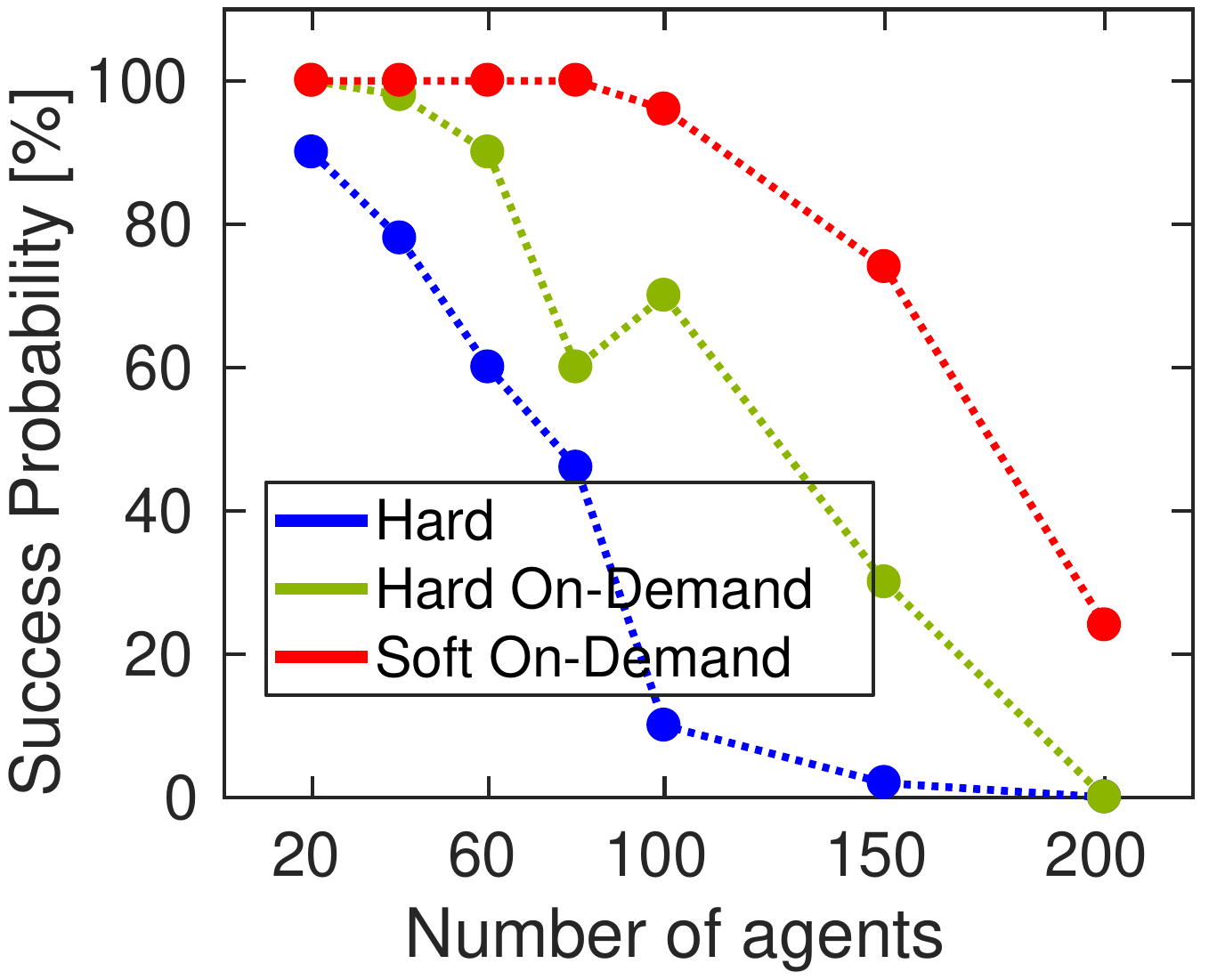}}
	\subfloat[]{\label{fig:time_hardsoft}\includegraphics[width=0.25\textwidth]{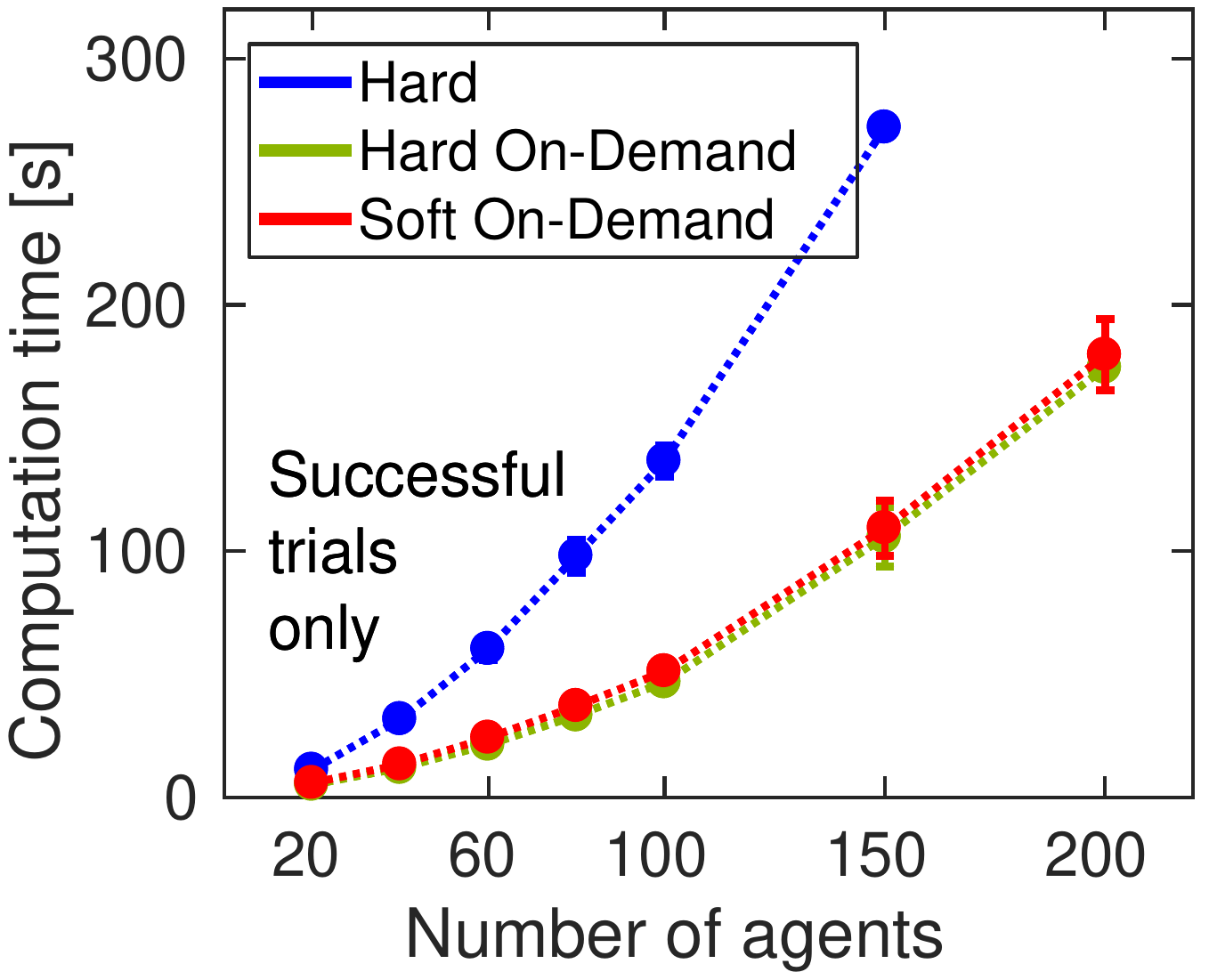}}
	\hfill
	\caption{\todo{Performance comparison of different collision avoidance strategies in DMPC, for an increasing number of agents within a workspace with a fixed agent density of $1 \hspace{1ex} \text{agent/m}^3$. For every swarm size considered, 50 different random test cases were generated.}}
	\label{fig:hardsoft}
\end{figure}

\begin{figure}[t]
	\vspace{0.5ex}
	\centering
	\subfloat[]{\label{fig:comp_prob}\includegraphics[width=0.25\textwidth]{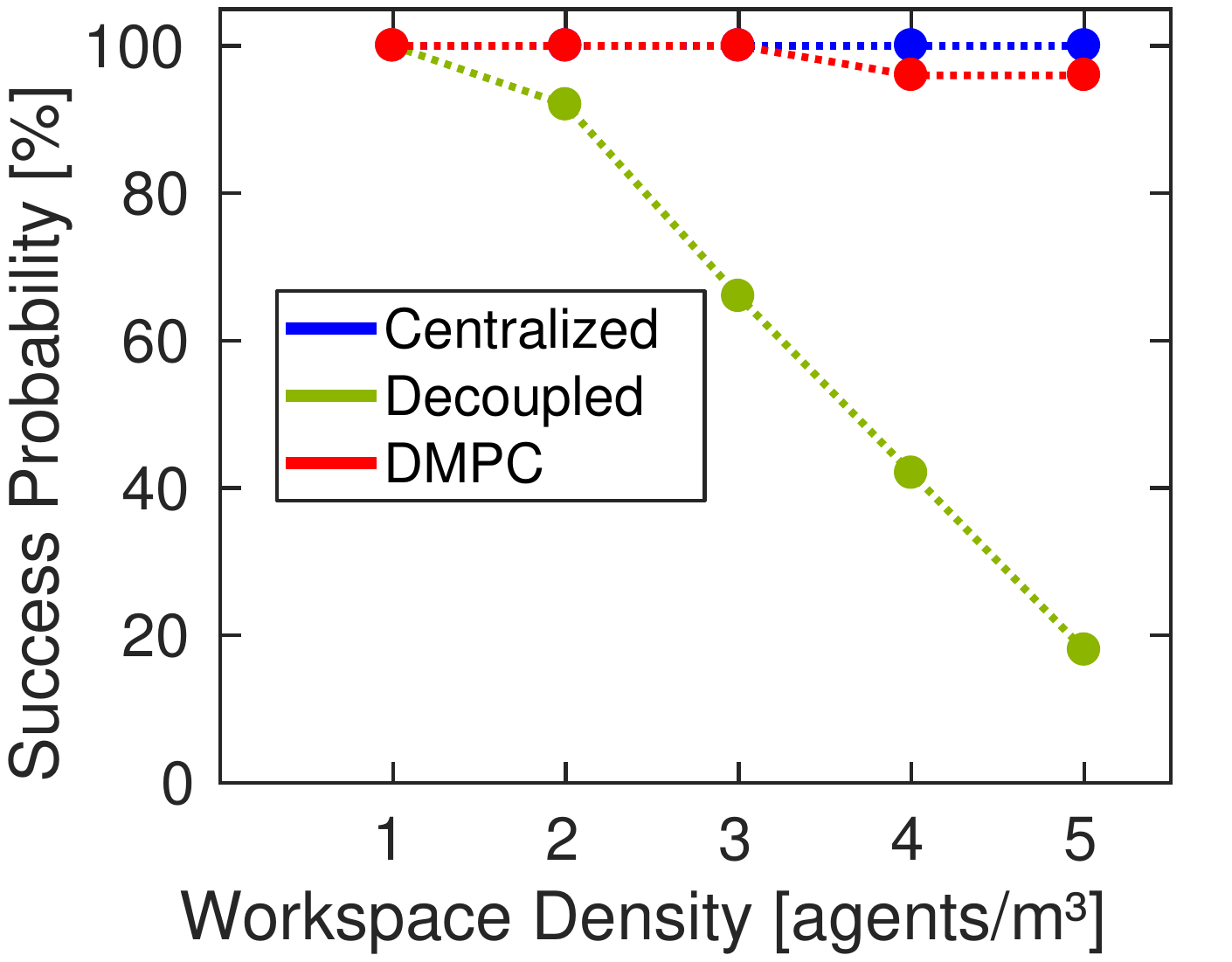}}
	\subfloat[]{\label{fig:comp_time}\includegraphics[width=0.25\textwidth]{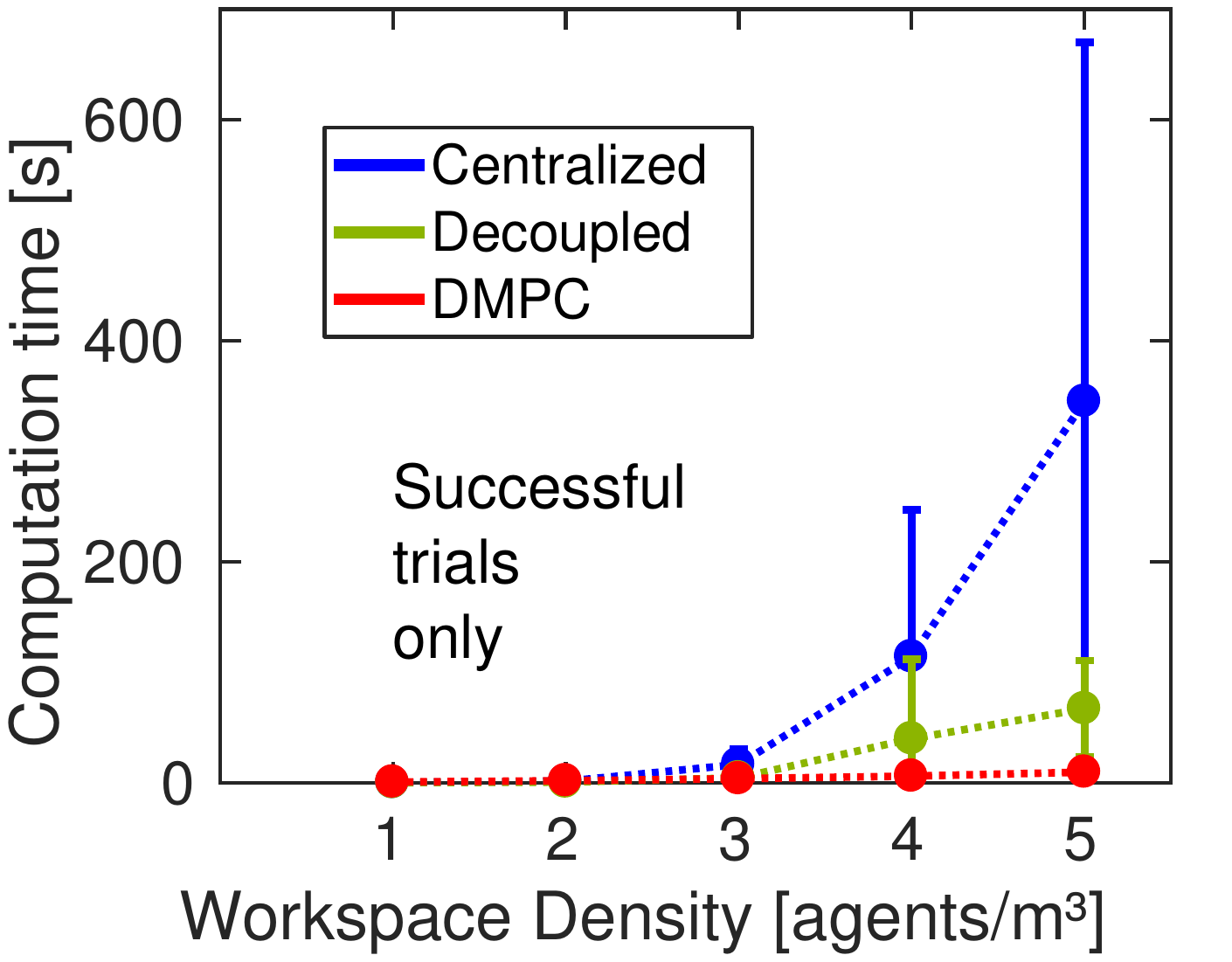}}
	\hfill
	\vspace{-2ex}
	\subfloat[]{\label{fig:comp_dist}\includegraphics[width=0.25\textwidth]{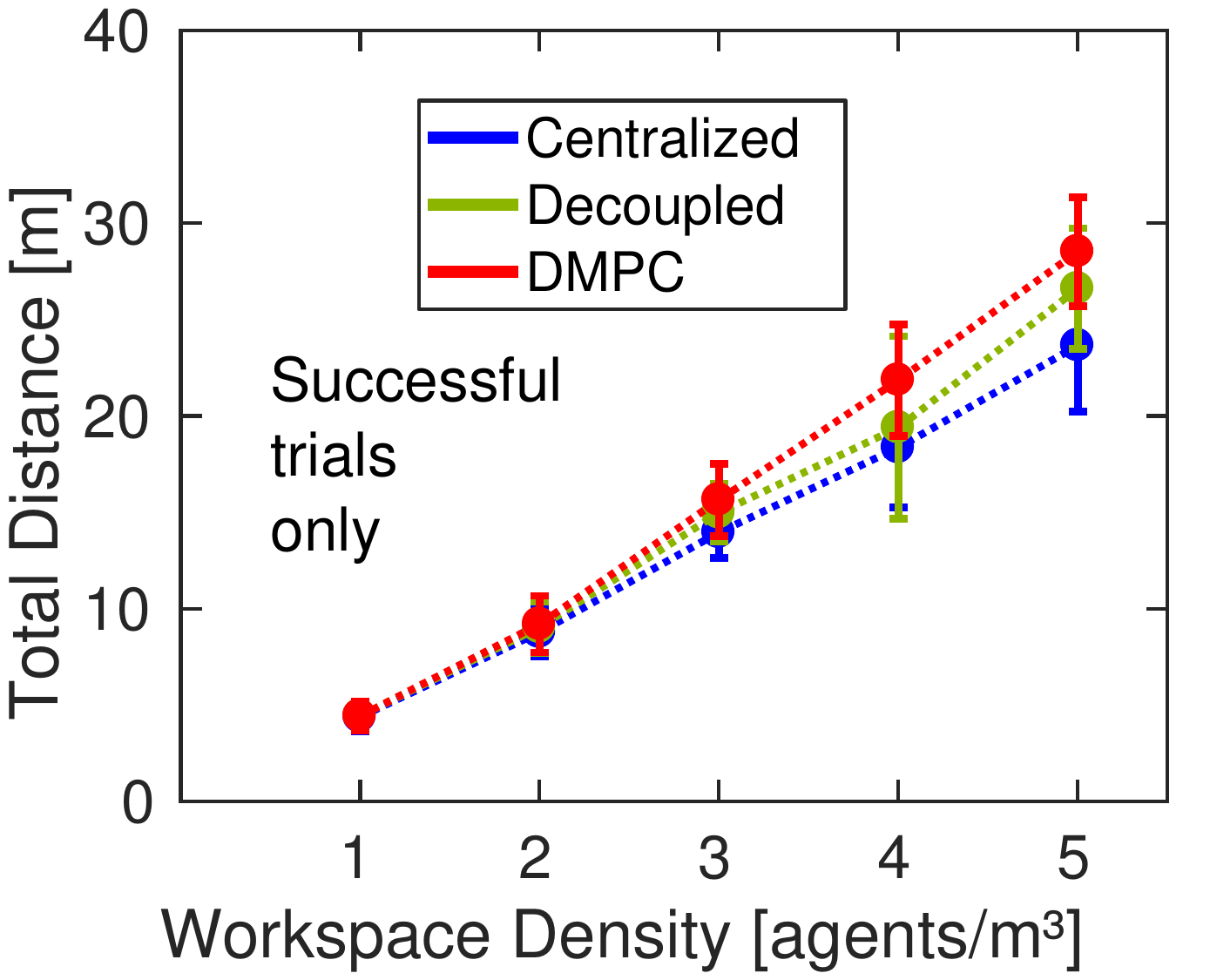}}
	\subfloat[]{\label{fig:traj_time}\includegraphics[width=0.25\textwidth]{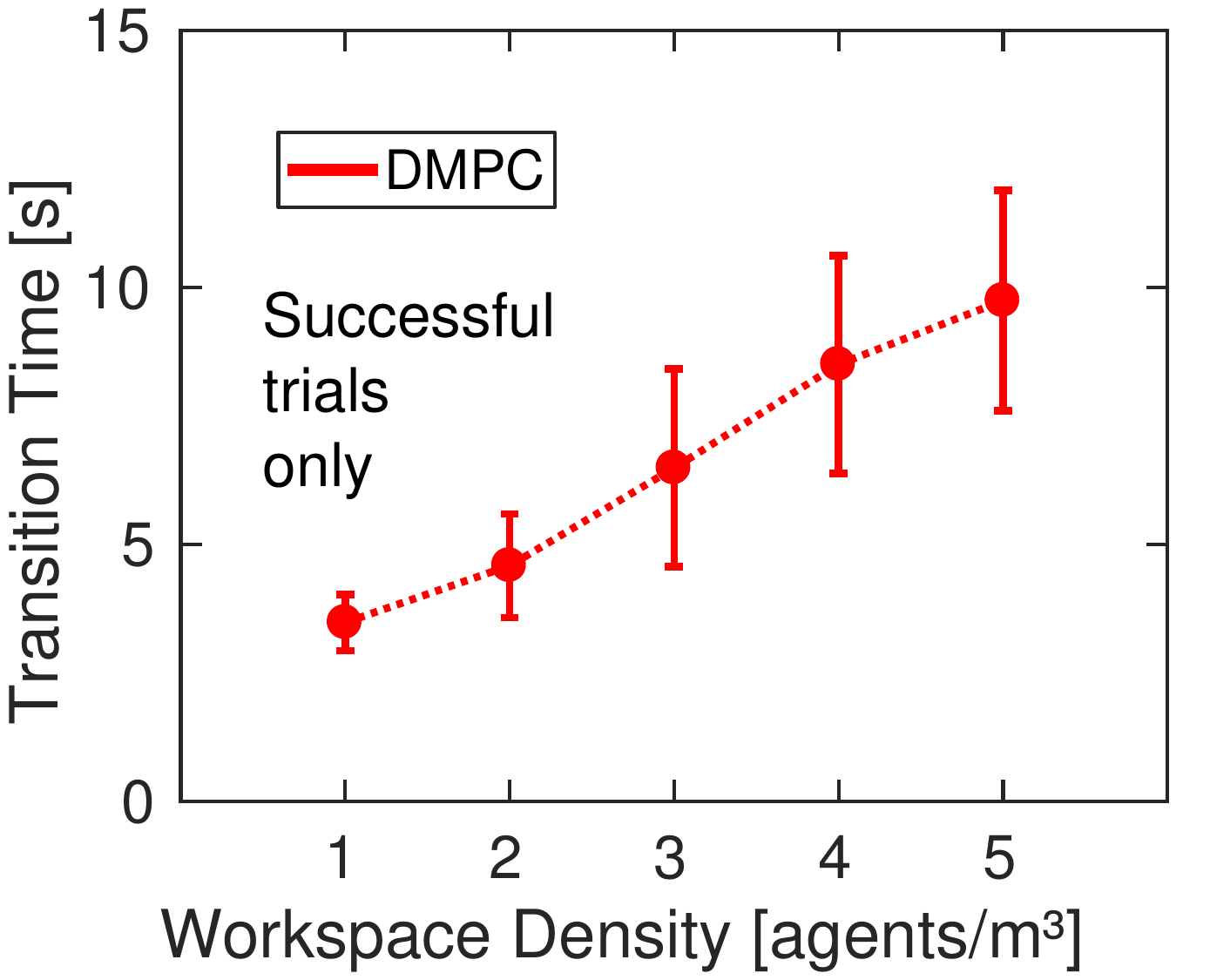}}

	\caption{Performance comparison of DMPC against SCP-based approaches, in a fixed $4 \text{m}^3$ volume. For every density considered, 50 different random test cases were generated.}
	\label{fig:sim}
\end{figure}

\subsection{\todo{Comparison of Collision Avoidance Strategies in DMPC}}
\label{subsec:hardvssoft}
\todo{To validate our on-demand collision avoidance scheme with soft constraints, we compared the performance to two other methods: (1) using hard collision constraints in every time step of the horizon (as in \cite{van2017distributed}) and (2) implementing our on-demand collision avoidance with hard constraints (i.e., constraint (\ref{eqn:coll_ellip}) without the relaxation variable). All methods were tested in scenarios with random sets of initial and final positions. We kept the density of the workspace (defined as $\text{agent/m}^3$) constant and varied the amount of agents from 20 to 200. All three approaches shared the time step parameters ${h = 0.2\,\text{s}}$ and $T_s = 0.01\,\text{s}$. We used a horizon length $K=15$, parameter $\kappa = 1$, a maximum relaxation of $\varepsilon_{\max} = 0.05\,\text{m}$ for the optimizer, a maximum relaxation of $\varepsilon_{\text{check}} = 0.05\,\text{m}$ for the safety check  and a maximum time to complete the transition $T_{\max} = 20\,\text{s}$. Fig.~\ref{fig:prob_hardsoft} shows the success rate of  DMPC for point-to-point transitions using the different collision avoidance schemes. If we use hard constraints at every time step (blue lines), the success rate suffers due to the inability of the agents to arrive at their final locations. The agents display conservative behaviour to maintain collision-free updates along their predictions, which may preclude progress towards the goal. On the other hand, the use of on-demand collision avoidance with hard constraints may lead to infeasible optimization problems, since the agents may be unable to avoid collisions within their acceleration limits. Our soft constraint strategy resolves the problem and achieves more than 75\% success rate with up to 150 agents, clearly outperforming the other two methods. The decrease in success rate for  200 agents is partially due to insufficient time to complete the transition leading to 55\% of the failures; with more agents and a fixed agent density (i.e., a larger environment) the average time to complete a random transition increases. This may mean that 55\% of the transitions are infeasible independent of the algorithm used. In addition, the introduction of more decision-making agents leads to more collisions (45\% of the failures).  In Fig.~\ref{fig:time_hardsoft} we highlight the reduction in computation time with our on-demand collision avoidance strategy.  }

\subsection{Comparison to SCP-Based Approaches}
\label{subsec:scp}
We compared the performance of our proposed DMPC scheme with two state-of-the-art algorithms: centralized \cite{augugliaro2012generation} and decoupled \cite{chen2015decoupled} SCP. We used the same simulation parameters as in Sec.~\ref{subsec:hardvssoft}, but the volume of the workspace was kept fixed at $4\,\text{m}^3$, and the number of agents ranged from 4 to 20. \todo{We increased the value of $\kappa$ to $2$ to encourage agents to move to their goals, which showed better performance for high-density environments}. Since the centralized and decoupled approaches require a fixed arrival time, we first solved each test using DMPC and determined the required time to complete the transition, and then set that as the arrival time of the SCP methods. \todo{Similar results were obtained by setting a fixed arrival time for the SCP methods for every trial (i.e., not based on the DMPC completion time), and are omitted}. If DMPC failed to solve, the arrival time was set to $T_{\max} = 20\,\text{s}$. \todo{Both SCP methods were executed until convergence was achieved or the problem was deemed infeasible. }


Fig.~\ref{fig:comp_prob} shows the probability of success as the density of agents increases. The proposed DMPC algorithm was able to find a solution in more than 95\% of the trials, for every density scenario considered. The centralized approach was able to find a solution in every case, while the decoupled approach failed increasingly with increasing density.

As for the computation time, Fig.~\ref{fig:comp_time} shows a reduction of up to 97\% in computation time with respect to centralized SCP and of 85\% with decoupled SCP. The runtime variance observed in the other two approaches is due to the test-by-test variance in arrival time, as seen in Fig.~\ref{fig:traj_time}. Note that this DMPC implementation does not exploit the parallelizable nature of the algorithm yet and already achieves significantly lower runtimes.

To measure the optimality of the generated trajectories we analysed the sum of travelled distances by the agents, as highlighted in Fig.~\ref{fig:comp_dist}. Our distributed approach produces longer paths on average, with respect to both the centralized and decoupled SCP. The suboptimality increases with workspace density, since the agents actively adjust their trajectories to avoid collisions, and oftentimes those adjustments lead  to non-optimal paths towards their goals.



\begin{figure}[t]
	\vspace{1ex}
	\centering
	\includegraphics[width=0.25\textwidth]{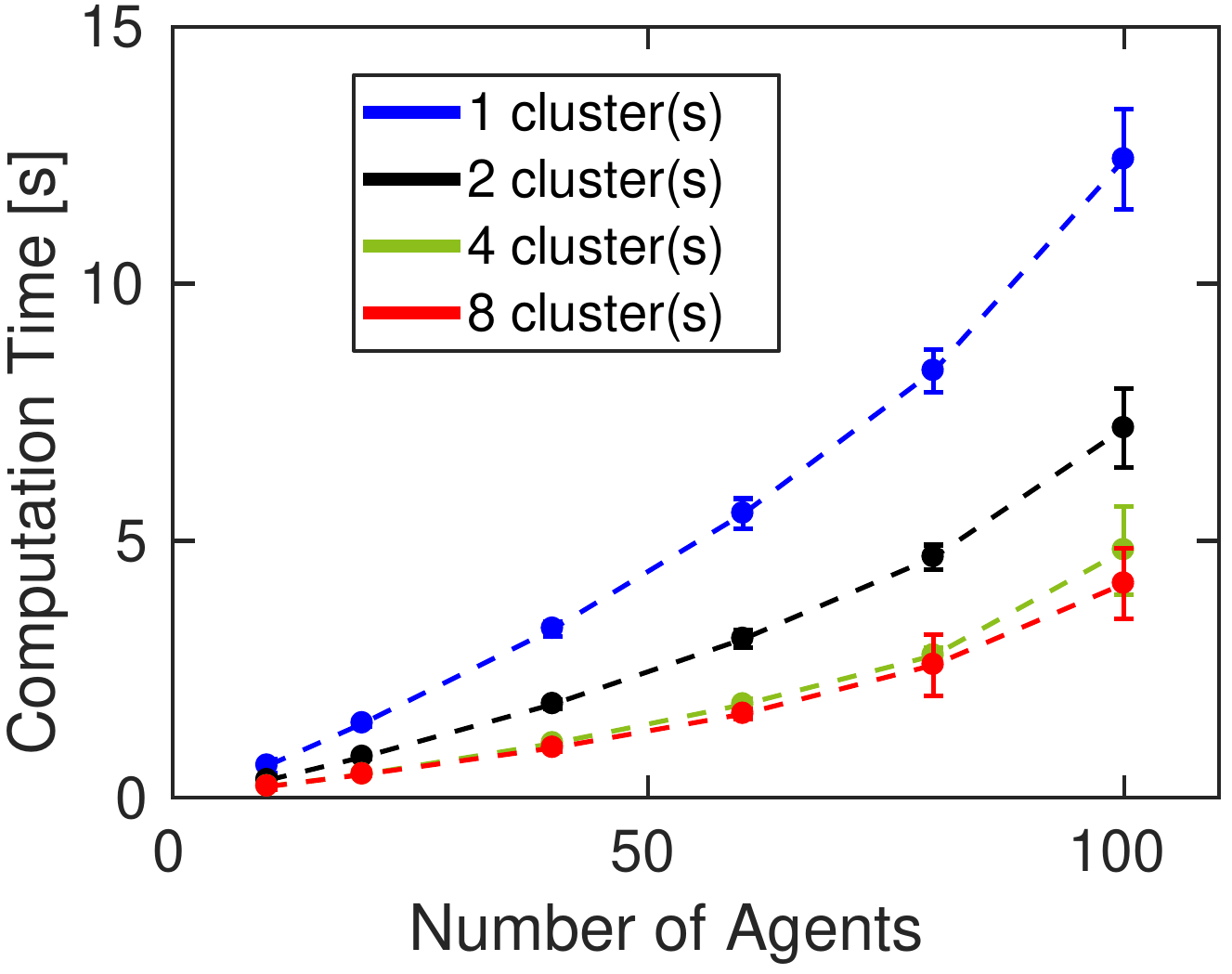}
	\caption{Average computation time for different numbers of clusters. For each swarm size, we gathered data of 30 successful transitions and reported the mean and standard deviation (vertical bars) of the runtime.}
	\label{fig:comp_clusters}
\end{figure}

\section{Experiments}
\label{sec:exp}
In this section we present experimental results using Algorithm 1 as an offline trajectory planner for a swarm of Crazyflies 2.0. The algorithm was implemented in C++ using OOQP as the solver. A video of the performance is found at \href{http://tiny.cc/dmpc-swarm}{{\tt http://tiny.cc/dmpc-swarm}}.

\subsection{Parallel DMPC}

Leveraging the parallel nature of the inner loop of Algorithm~1, we can design a strategy that parallelizes the computation. The idea is to equally split the $N$ agents into smaller clusters to be solved in parallel using a multicore processor. The optimization problems of the agents inside a cluster are solved sequentially, but with the advantage of iterating through fewer agents. After all the clusters finish solving their QPs, they exchange the updated predictions and repeat the process.

In Fig.~\ref{fig:comp_clusters} we compare different numbers of clusters tested on a wide variety of transition scenarios. It was found that 8 clusters led to the best result for our computing hardware (CPU with 8 cores). This parallel strategy (8 clusters) reduced the computation time by more than 60\% compared to using a purely sequential execution (1 cluster).

\subsection{Swarm Transition}
\begin{figure}
	\centering
	\subfloat[]{\label{fig:grid}\includegraphics[width=0.23\textwidth]{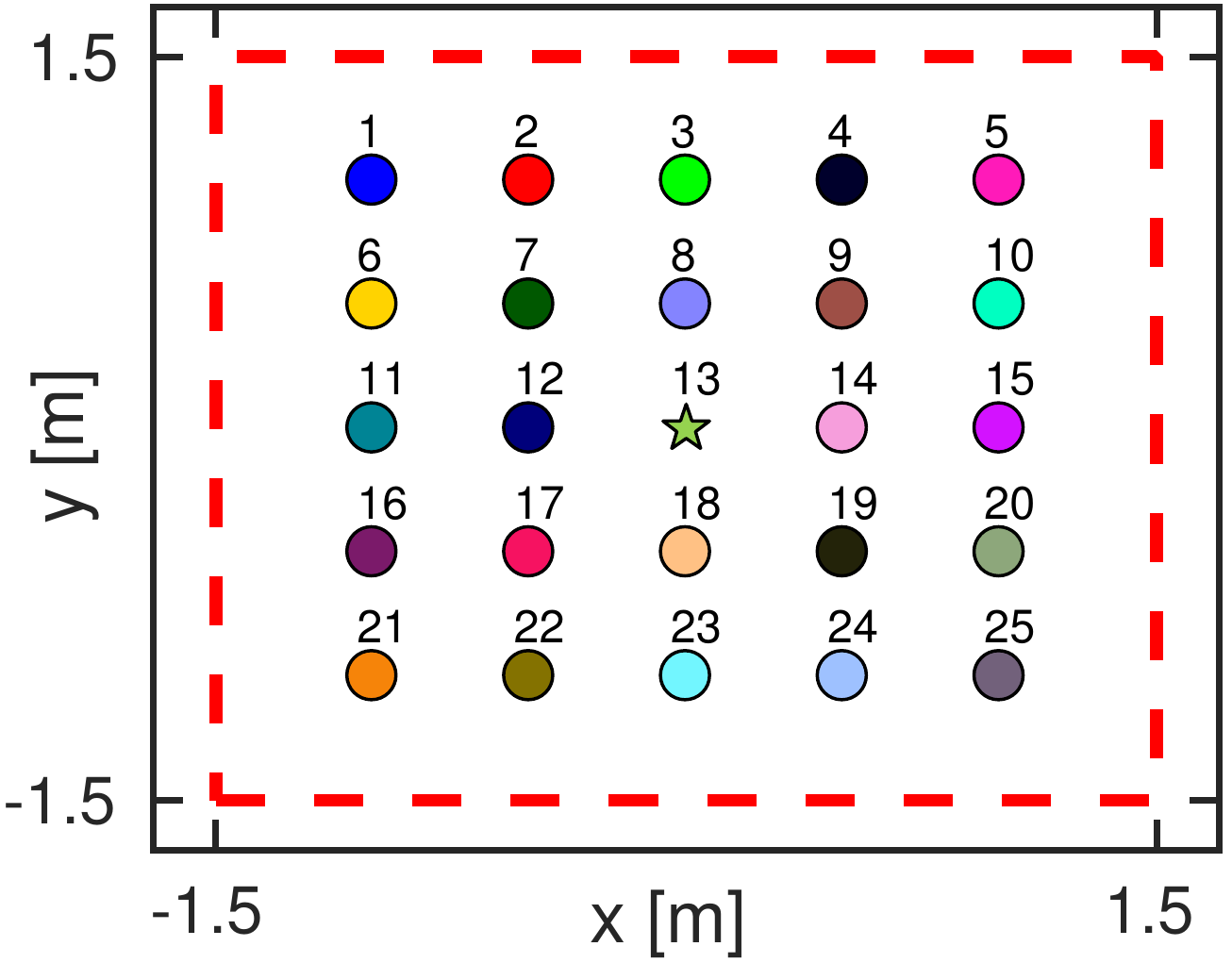}}
	\hfill
	\subfloat[]{\label{fig:dsl}\includegraphics[width=0.23\textwidth]{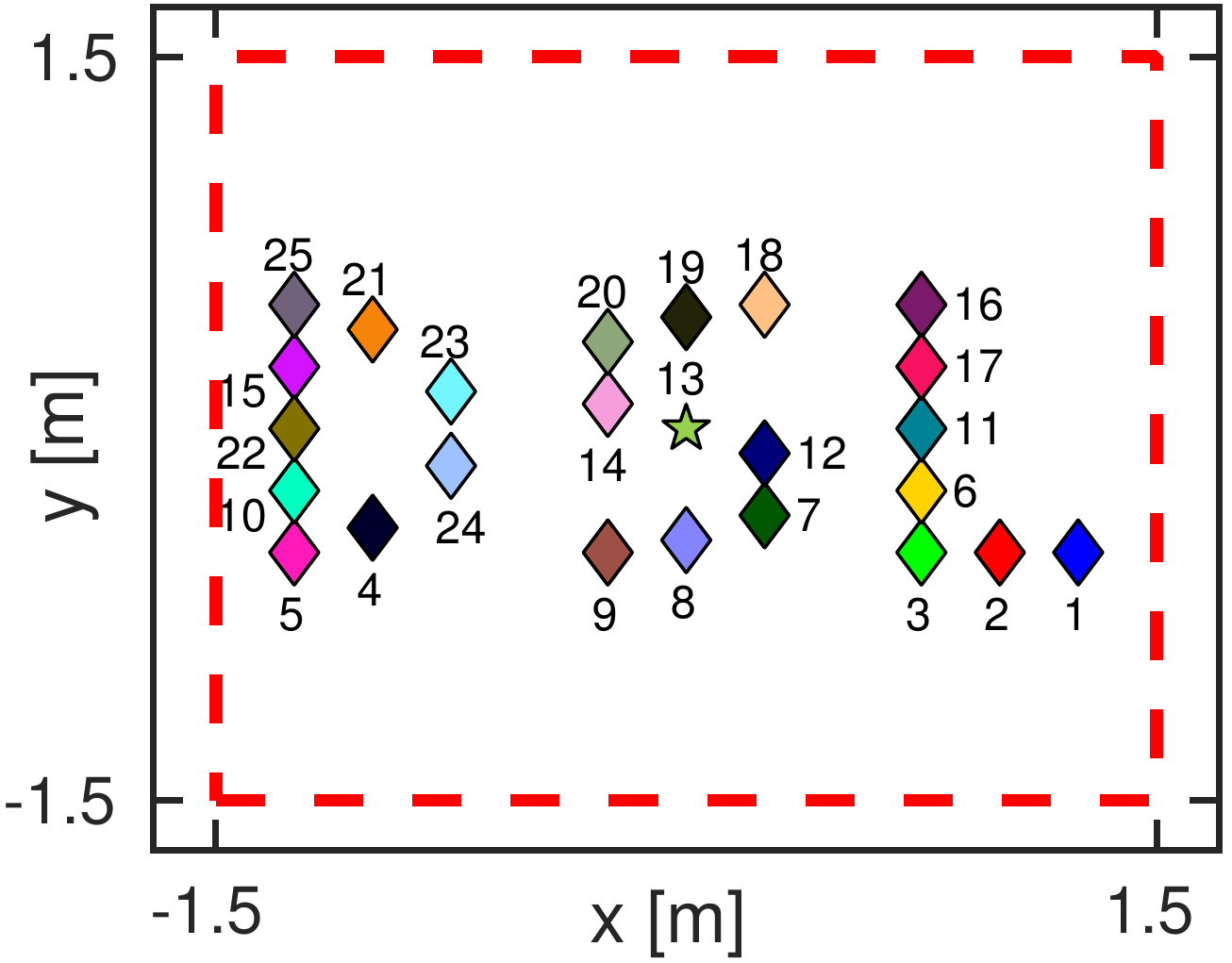}}
	

	
	%
	\caption{A 25-agent transition scenario: \protect\subref{fig:grid} initial grid configuration, \protect\subref{fig:dsl} target `DSL' configuration. Circles and diamonds (of matching colour) represent initial and final locations for all agents, respectively. The star in the middle represents an agent acting as a static obstacle. The bounding box in dashed red lines represents the workspace boundaries.  }
	\label{fig:experimental} 
\end{figure}

To perform the pre-computed transition motion on the quadrotors, we communicated via radio link with each drone and sent the following information at 100 Hz: (1) position setpoints and (2) position estimates from an overhead motion capture system. The setpoints were tracked using an on-board position controller based on \cite{mellinger2011minimum}. One transition scenario is depicted in Fig.~\ref{fig:experimental}, in which the swarm was to transition from a $5\times 5$ grid to a `DSL' configuration. The difficulty of this particular scenario was increased by the central agent acting as a static obstacle (i.e., obstacle with fixed position). 

\begin{figure}
	\vspace{1ex}
	\centering
	\subfloat[]{\label{fig:exp_dist}\includegraphics[trim = {0.8cm 0.2cm 2.0cm 0.5cm},clip,width=0.42\textwidth]{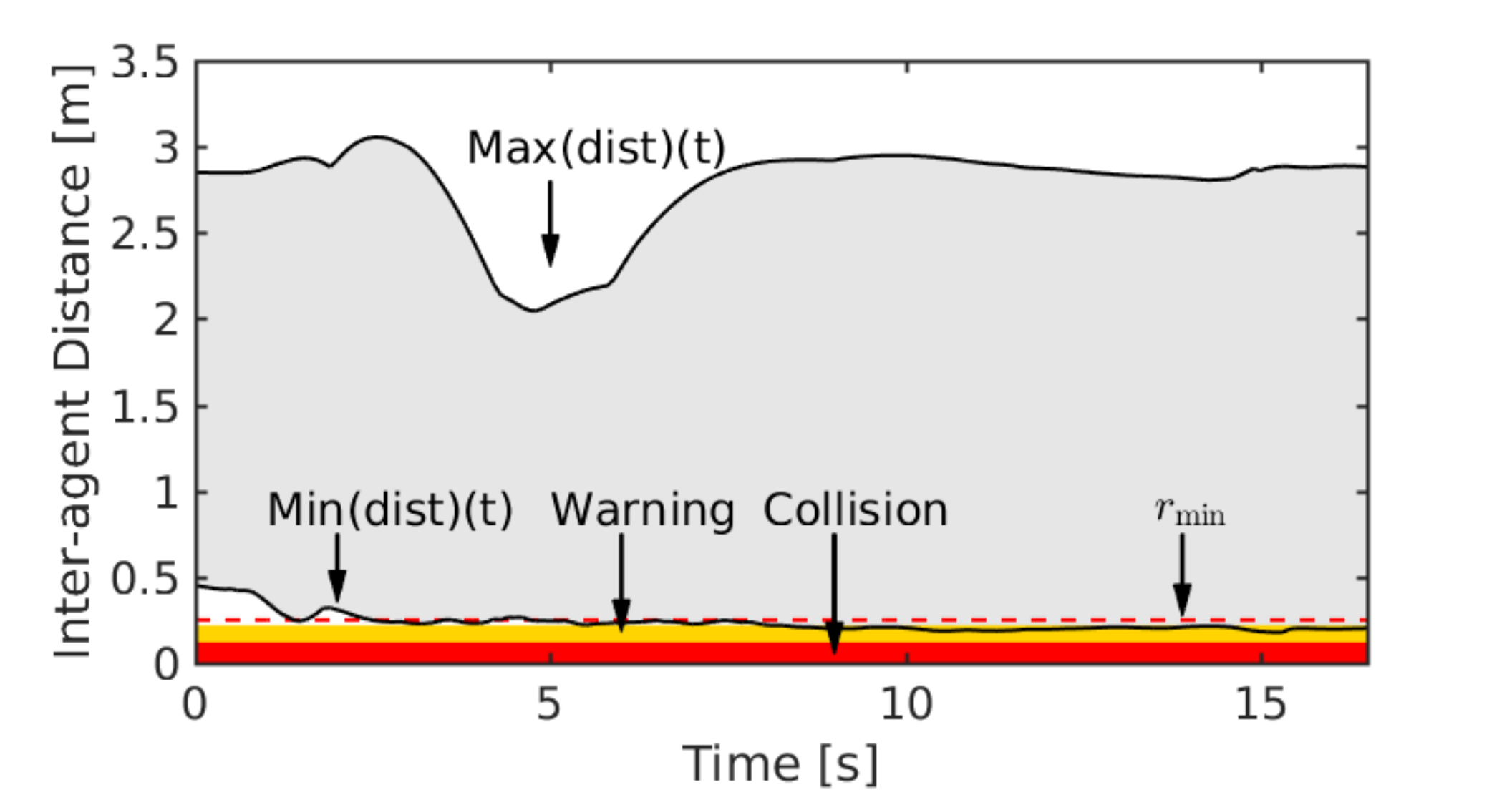}}
	\hfill
	\subfloat[]{\label{fig:exp_goal}\includegraphics[trim = {0cm 0cm 0cm 0cm},clip,width=0.42\textwidth]{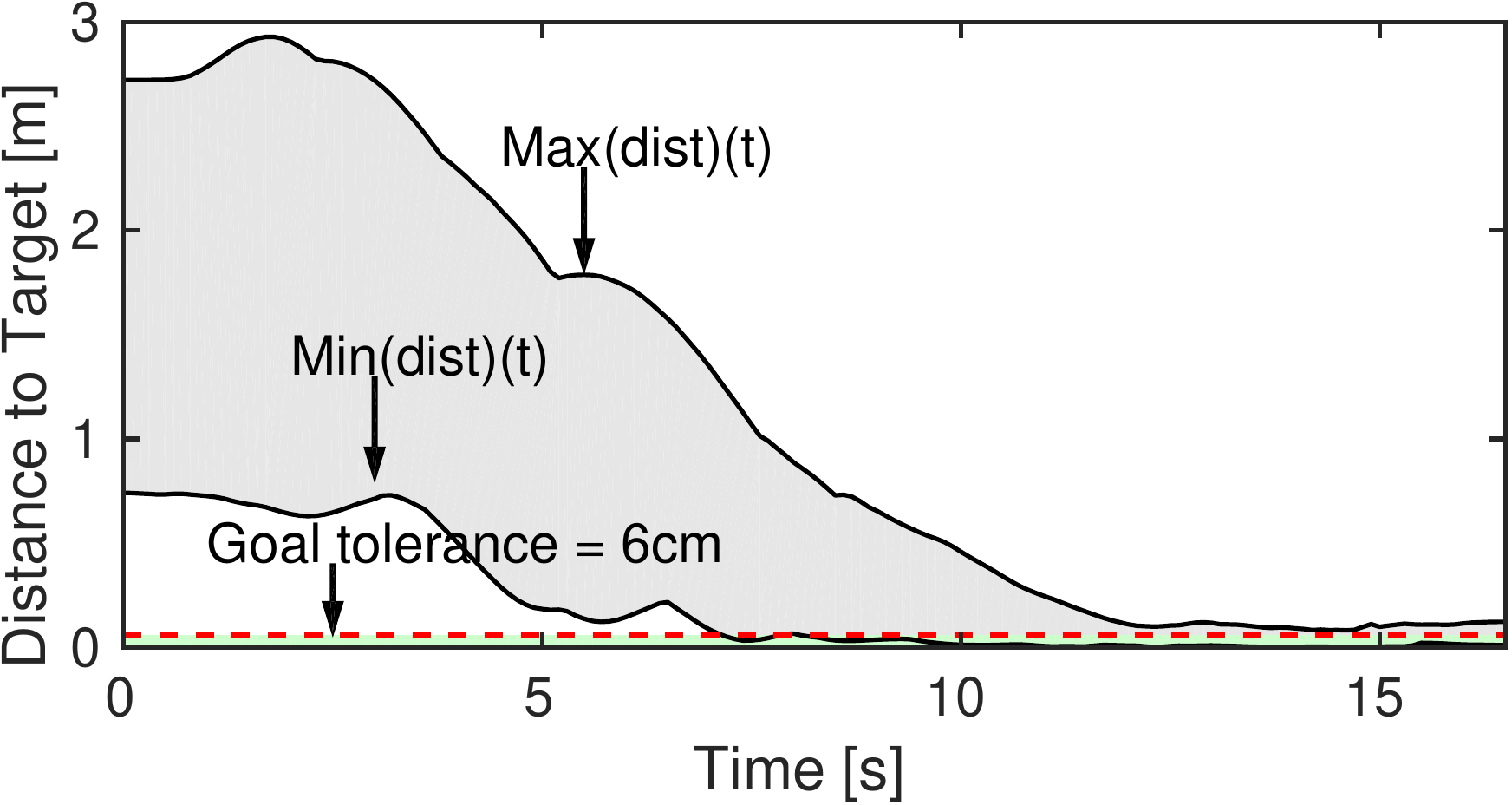}}
	\hfill

	\caption{Experimental data from the transition depicted in Fig.~\ref{fig:experimental}, showing maximum and minimum distance values over 6 independent trials: \protect\subref{fig:exp_dist}~pairwise distances, \protect\subref{fig:exp_goal} distances to target locations.}
	\label{fig:exp} 
\end{figure}

\todo{We required $r_{\min} = 0.25\,\text{m}$ with $\varepsilon_{\text{check}} = 0.03\,\text{m}$}. The DMPC algorithm was able to find a solution for this scenario in 1.8 seconds. In Fig.~\ref{fig:exp_dist}, the curves delimiting the gray area correspond to the minimum and maximum inter-agent distance at each time instant for six independent executions of the transition. Although trajectories are planned such that any inter-agent distance must remain above the warning zone (yellow band), the experimental curve goes slightly below that value. The warning zone is, in practice, a safety margin to compensate for unmodelled phenomena in our planning algorithm, such as imperfect trajectory tracking, time delays, and aerodynamics. Taking all these factors into account, it is natural for the minimum distance curve to go farther below than planned; however, it still remains above the collision zone. It is critical for the warning zone to be large enough, as to absorb any mismatch between the idealized planning and the real world. Its size is directly controlled by $r_{\min}$, which must be carefully chosen for robust trajectory executions.


Finally, Fig.~\ref{fig:exp_goal} shows that the agents' progress towards their goal and are able to complete the transition up to some small tolerance. Once the agents enter the tolerance region below the dashed red line, they were commanded to hover in place. The on-board position controller reported a maximum error of close to 3 cm during hover, which explains why the maximum distance curve remains slightly above the tolerance region after all agents reached their goals. 

In addition to the showcased scenario, the system has been tested on many randomly generated transitions, as can be seen in the video that accompanies this paper.

\section{Conclusions}
\label{sec:conclusions}

The DMPC algorithm developed in this paper enables fast multiagent point-to-point trajectory generation. Using model-based predictions, the agents detect and avoid future collisions while moving to their goal locations. \todo{We introduced on-demand collision avoidance with soft constraints in a DMPC framework to enhance the scalability and success rate over previous approaches. As compared to SCP-based methods, we drastically reduce computational complexity, with only a small impact on the optimality of the plans. Our formulation allows for parallel computing, which further reduces the runtime.}

\todo{We validated our method through an extensive empirical analysis using randomly generated transition tasks. Experimental results further validate our approach, which can be used to quickly calculate and execute transition trajectories for large teams of quadrotors, enabling new capabilities in applications such as drone shows.}

\bibliographystyle{IEEEtran}
\bibliography{IEEEabrv,reference}

\end{document}